%% file: neurips_2025.tex
\documentclass{article}

\usepackage{graphicx}

\usepackage[final]{neurips_2025}

\usepackage[utf8]{inputenc} % allow utf-8 input
\usepackage[T1]{fontenc}    % use 8-bit T1 fonts
\usepackage{hyperref}       % hyperlinks
\usepackage{url}            % simple URL typesetting
\usepackage{booktabs}       % professional-quality tables
\usepackage{amsfonts}       % blackboard math symbols
\usepackage{nicefrac}       % compact symbols for 1/2, etc.
\usepackage{microtype}      % microtypography
\usepackage{xcolor}         % colors
\usepackage{amsmath}
\usepackage{pifont}
\usepackage{tabularx}
\usepackage{multicol}
\usepackage{multirow}
\usepackage{float}
\usepackage{wrapfig}
\usepackage{subcaption}

\usepackage{cleveref}
\crefname{section}{§}{§§}
\Crefname{section}{§}{§§}

\usepackage{colortbl}

\definecolor{mygray}{rgb}{0.95,0.95,0.95}

\title{
State Space Prompting via Gathering and Spreading Spatio-Temporal Information for Video Understanding
}

\author{%
  Jiahuan Zhou$^1$, Kai Zhu$^1$, Zhenyu Cui$^1$, Zichen Liu$^1$, Xu Zou$^2$\thanks{Corresponding author}, Gang Hua$^3$\\
   $^1$Wangxuan Institute of Computer Technology, Peking University, Beijing 100871, China\\
  $^2$the Huazhong University of Science and Technology, Wuhan 430074,China\\
  $^3$Amazon.com, Inc, Bellevue, WA 98004, USA\\
  \texttt{jiahuanzhou@pku.edu.cn}, \texttt{zhukai2022@ruc.edu.cn} \\
  \texttt{\{cuizhenyu,lzc20180720\}@stu.pku.edu.cn}, \texttt{zx@zoux.me}, \texttt{ganghua@gmail.com}
}

\begin{document}

\maketitle

\begin{abstract}
Recently, pre-trained state space models have shown great potential for video classification, which sequentially compresses visual tokens in videos with linear complexity, thereby improving the processing efficiency of video data while maintaining high performance. To apply powerful pre-trained models to downstream tasks, prompt learning is proposed to achieve efficient downstream task adaptation with only a small number of fine-tuned parameters. However, the sequentially compressed visual prompt tokens fail to capture the spatial and temporal contextual information in the video, thus limiting the effective propagation of spatial information within a video frame and temporal information between frames in the state compression model and the extraction of discriminative information. To tackle the above issue, we proposed a State Space Prompting (SSP) method for video understanding, which combines intra-frame and inter-frame prompts to aggregate and propagate key spatiotemporal information in the video. Specifically, an Intra-Frame Gathering (IFG) module is designed to aggregate spatial key information within each frame. Besides, an Inter-Frame Spreading (IFS) module is designed to spread discriminative spatio-temporal information across different frames. By adaptively balancing and compressing key spatio-temporal information within and between frames, our SSP effectively propagates discriminative information in videos in a complementary manner. Extensive experiments on four video benchmark datasets verify that our SSP significantly outperforms existing SOTA methods by 2.76\% on average while reducing the overhead of fine-tuning parameters. The code is available at \href{https://github.com/zhoujiahuan1991/NeurIPS2025-SSP}{https://github.com/zhoujiahuan1991/NeurIPS2025-SSP}.
\end{abstract}

\section{Introduction}

In recent years, the Vision Transformer (ViT) has demonstrated its promising performance in video understanding due to its powerful attention-based context modelling capabilities~\cite{ViT, vivit, timesformer, zhou2025distribution, xu2025self, xu2025long, xu2025dask, xu2024mitigate, xu2024lstkc, xu2024distribution}. However, the computational cost of the attention mechanism, which increases quadratically with the length of the input data, incurs huge computational and memory costs, especially when processing long video sequences. To achieve efficient video processing, a state space modelling method, called VideoMamba, is introduced to achieve comparable performance to the ViT while maintaining linear computational complexity~\cite{park2024videomamba, liu2022video}. Despite some progress in model pre-training\cite{videomae, videomae_v2, mamba_pretrain}, it still suffers from the heavy overhead of downstream task adaptation through parameter fine-tuning. Therefore, Parameter-Efficient Fine-Tuning (PEFT) has aroused extensive attention to achieve comparable or even higher performance than Full Fine-Tuning (FFT) and reduce the adaptation costs by optimizing only a few set of parameters~\cite{fu2023effectiveness, visual_tuning, peft_vision}.

% However, the quadratic time complexity of attention computation leads to substantial computational and memory overhead when processing long-context video sequences. With the advancement of state-space models, VideoMamba introduces an input-dependent selective state space modeling mechanism into video understanding, achieving performance comparable to or surpassing mainstream Transformer architectures while maintaining linear computational and memory complexity. This breakthrough positions VideoMamba as a promising next-generation backbone for video understanding. Nevertheless, directly fine-tuning large pre-trained video models on downstream tasks faces prohibitive computational costs due to increased model parameters, massive video datasets, and domain shifts. This challenge has motivated research on parameter-efficient fine-tuning methods that optimize only a subset of parameters or incorporate minimal learnable components to reduce computational and storage requirements during adaptation.

Early PEFT methods mainly focused on parameter-level efficient fine-tuning, \emph{e.g.,} Adapter~\cite{adapterformer, vl-adapter, st-adapter, adapterstrike} or LoRA~\cite{lora, mtlora}, but still suffered from inefficiency with the additional introduction of model parameters. To this end, visual prompt learning technology aims to embed a small set of learnable prompting tokens at the input level, enabling efficient downstream task adaptation without extending internal model parameters~\cite{vpt, vfpt, revisiting_vpt, e2vpt, autovp, insvp, CoOp, CoCoOp, xu2025componential, zhang2025scap}. However, as shown in Figure~\ref{fig: motivation}(a), existing video prompting methods are typically ViT-oriented, which exploit global attention mechanisms to propagate key information in intra- and inter-frame prompts~\cite{VoP, DGL, MPT, STOP}, ignoring the requirement to balance efficiency and effectiveness in video understanding for VideoMamba. Specifically, ignoring the sequential compression state space~\cite{S4, lssl, dss, gu2020hippo, gu2022parameterization}, existing video prompting methods fail to gather spatial information in long video token sequences for Mamba. In addition, the global attention-oriented prompting method mitigates the high efficiency in VideoMamba, leading to the dissemination of key spatial and temporal information in the long video token sequence. Consequently, as shown in Figure~\ref{fig: motivation}(b), the embedded prompts mitigate the aggregation of discriminative spatio-temporal information due to the information decay after long-term compression~\cite{longmamba2025longmamba}.

\vspace{5pt}

\noindent
\begin{minipage}{\textwidth}
\centering
\includegraphics[width=1.0\textwidth]{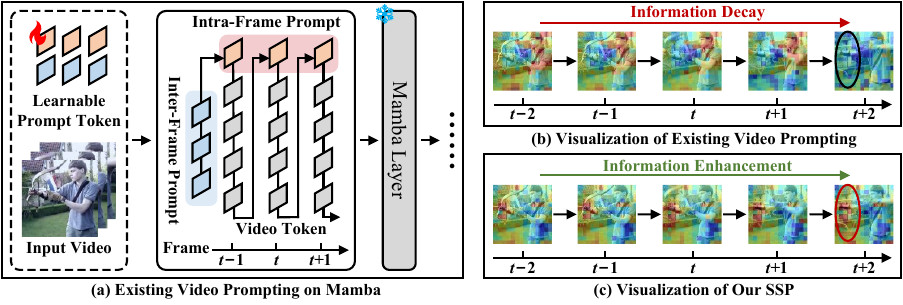}
\\[-3pt]
\captionof{figure}{Existing video prompting methods on Mamba and its visualization results of updating date compared to our SSP. Existing methods directly concatenate learnable prompts to video tokens, resulting in the information decay problem after the long-term state space compression. However, our SSP achieves information enhancement through spatial gathering and temporal spreading.}
\label{fig: motivation}
\end{minipage}

\vspace{7pt}

To address the above challenges, we propose a State Space Prompting (SSP) framework for video understanding, which gathers and spreads spatio-temporal information in an efficient and effective manner, as shown in Figure~\ref{fig: motivation}(c). Specifically, an Intra-Frame Gathering (IFG) module is designed to exploit a low-rank local convolution to aggregate spatial information within each video frame. Sequentially, a low-rank attention-oriented Inter-Frame Spreading (IFS) module is further proposed to spread key information at the temporal level, which develops a low-rank attention module to refine the temporal information that is gradually aggregated between frames, where long-term context information is effectively spread within global temporal prompts. Among them, the information entropy of each frame adjusts the attention given to each frame when refining the temporal information, and the frame-specific spatial variance is employed to gate the influence strength from the spreading information to each frame.

% To address these challenges, we propose a novel and effective dual-stream prompting framework: (1) we sample the last token of each frame along Mamba's scanning trajectory to gather hidden state information from all preceding frames. These features are then spread into multiple parallel inter-frame prompt generation modules for low-rank attention computation, producing inter-frame prompts that are inserted at token boundaries between frames. This design ensures that prompts effectively gather and spread spatio-tenporal information from hidden states, enabling the Mamba model to leverage comprehensive global temporal cues during token processing. (2) We develop a lightweight intra-frame prompt generation module that employs low-rank 2D convolutions on per-frame tokens to create intra-frame prompts, which are superimposed on input data. These intra-frame prompts further refine inter-frame operations through adaptive weight adjustment and gating mechanisms applied to the module's inputs and outputs. \\

In summary, our contributions are three-fold: (1) We proposed a State Space Prompting method for Video Understanding, called SSP, which gathers and spreads discriminative spatio-temporal information compressed by the state space model to achieve high effectiveness while maintaining its computational efficiency. (2) We design an Intra-Frame Gathering module and an Inter-Frame Spreading module to facilitate spatio-temporal contextual information interaction by spreading gathered local spatial information in a temporal manner. (3) Extensive experiments on multiple video understanding benchmarks demonstrate that our method achieves superior performance against existing methods with only $\sim$3\% of tunable parameters compared to full tuning.

\section{Related Work}

\subsection{State Space Model (SSM)}

In recent years, State Space Models (SSMs) have emerged as a promising approach for sequence modeling, offering the ability to capture long-range dependencies with linear computational complexity~\cite{S4, lssl, dss, gu2020hippo, gu2022parameterization}. Building on this foundation, Mamba and Mamba2 introduced input-dependent update and forget gates to address limitations in content-based reasoning~\cite{mamba, mamba2}. Unlike Transformer architectures, this allows SSM parameters to be dynamically modulated by input, significantly improving expressiveness in discrete modalities. Coupled with hardware-friendly parallelization, these advances lead to notable gains in computational efficiency and performance in language modeling.

Building on this progress, Vision Mamba extended the Mamba framework to 2D image modeling via bidirectional spatial scanning, achieving strong results in image understanding tasks~\cite{vim, liu2024vmamba}. More recently, VideoMamba further generalized this approach to video by introducing spatio-temporal scanning, enabling efficient modeling of global dependencies across both spatial and temporal dimensions with linear complexity~\cite{park2024videomamba, li2024videomamba}. This architecture rivals or surpasses traditional CNN and Transformer models~\cite{liu2022video, feichtenhofer2019slowfast, feichtenhofer2020x3d, li2022uniformer}, establishing itself as a competitive backbone for video understanding. However, applying pre-trained models like VideoMamba to downstream tasks using full fine-tuning typically requires large data and high computational cost. Thus, developing parameter-efficient fine-tuning strategies for VideoMamba remains an important and urgent challenge.

\subsection{Parameter-Efficient Fine-Tuning for SSM}

Parameter-efficient fine-tuning techniques aim to reduce learnable parameters while maintaining model performance, thereby reducing storage and computational costs when adapting pre-trained models to downstream tasks~\cite{fu2023effectiveness}. Several studies have attempted to apply parameter-efficient fine-tuning methods to Mamba architecture models~\cite{svp, ham2024parameter, yoshimura2024mambapeft}. These methods can be categorized into partial-based, addition-based, and prompt-based approaches.

\textbf{Partial-based methods} typically fine-tune only a subset of parameters in the pre-trained Mamba model, such as projection layers, convolution layers, or forget gates~\cite{ham2024parameter, yoshimura2024mambapeft}. These methods are straightforward and easy to implement. However, partial-based approaches are constrained by the model's inherent parameter space, limiting their adaptability to downstream tasks across different domains.
\textbf{Addition-based methods} generally freeze the original model parameters and incorporate learnable components, such as adapter modules~\cite{adapterformer, st-adapter}. While these plug-and-play modules can be readily transferred to the Mamba model architecture, they simply apply transformations to the input data without considering the sequential progression characteristics inherent to Mamba's architecture. The Additional-Scan approach attempts to learn downstream knowledge by increasing the state dimensions of the SSM~\cite{yoshimura2024mambapeft}. However, when applied to video domains, merely increasing state dimensions proves insufficient for effectively extracting critical spatio-temporal information.
\textbf{Prompt-based methods} add a small number of learnable prompt vectors, optimizing only these parameters during training. Existing work such as SVP has transferred prompt learning methods to Mamba models by generating prompts for each token, effectively activating Mamba's update and forget gates during fine-tuning~\cite{svp}. However, such methods are designed exclusively for static 2D images. When transferred to video tasks, they similarly struggle with the challenge of modeling long-context spatio-temporal relationships.

\subsection{Video Prompting}

Prompt learning methods were first introduced in natural language processing (NLP) to transfer pre-trained models to various downstream tasks~\cite{prompt, prefix-tuning, p-tuning, aprompt}. Inspired by the success of prompt learning in NLP, these methods have been extended to the visual domain\cite{li2024exemplar, liu2024compositional}. VPT and VFPT fine-tune models by concatenating token-based prompts with input data to capture image features~\cite{vpt, vfpt}. More recently, methods like DGL and STOP have applied prompt learning to video tasks, using intra-frame and inter-frame prompt modules to capture temporal information~\cite{VoP, DGL, MPT, STOP}. For instance, DGL employs prompt vectors as query, key and value vectors to model both local features and global features in videos~\cite{DGL}. MPT utilizes prompts as query vectors, leveraging the Q-Former mechanism to extract spatial, temporal, and global features~\cite{MPT}. STOP generates inter-frame prompts

\vspace{-5pt}

\begin{figure}[H]
  \centering
  \includegraphics[width=1\textwidth]{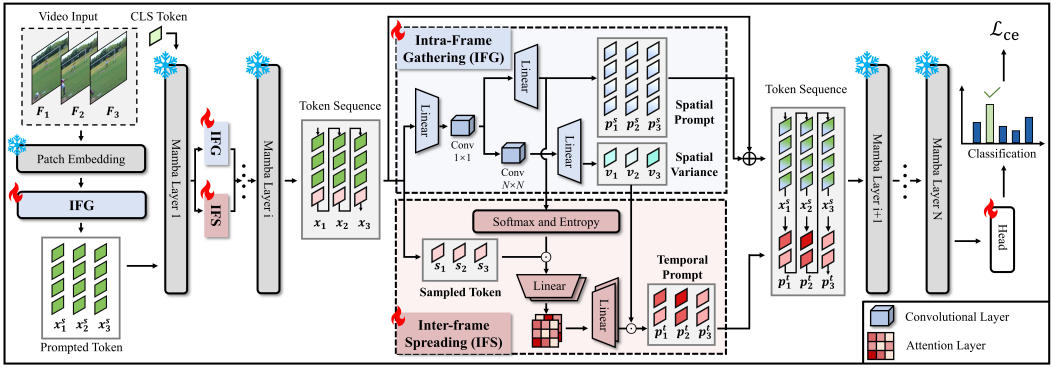}
  \vspace{-12pt}
  \caption{\footnotesize The pipeline of our SSP. We embed videos into image tokens with spatial prompts. After the initial Mamba layer, our complementary IFG and IFS modules operate - IFG aggregates spatial information while IFS spreads temporal information. Information entropy and frame-specific spatial variance bridge these modules. The prompted tokens and CLS token then pass through subsequent Mamba layers for video classification.}
  \label{fig:framework}
\end{figure}

\vspace{-15pt}

from all tokens while using intra-frame prompts to highlight the varying importance of frames, dynamically inserting prompts between frames~\cite{STOP}.

However, these methods were originally designed for the Transformer architecture, leveraging its unique structure by connecting prompts to query and key-value vectors to store task information. Transformer-based video prompting methods use a global attention mechanism~\cite{attention}, allowing tokens within the sequence to interact at any position. In contrast, the Mamba architecture propagates tokens sequentially~\cite{han2024demystify, rezaei2024mambalrp, zimerman2025explaining}, causing adjacent tokens to contain more overlapping information. As a result, when applying existing video prompting methods to Mamba, feeding all tokens into the inter-frame prompt module introduces redundant contextual information. This issue is more pronounced in video data with high spatiotemporal redundancy, making it difficult for the prompt module to effectively capture and propagate discriminative spatiotemporal context, ultimately limiting the model's performance.

%\label{gen_inst}

%\section{Citations, figures, tables, references}
%\label{others}

%\subsection{Citations within the text}

\section{Methodology}

In this section, we illustrate the proposed SSP comprehensively, and the overall pipeline is depicted in Figure \ref{fig:framework}.

% \subsection{Preliminary}
% \subsection{Preliminary of State Space Model}
\subsection{Preliminary of Mamba}

% \textbf{State Space Model.} 
The SSM-based models, Mamba, Vision Mamba (ViM), and VideoMamba, are inspired by continuous systems that map one-dimensional equations or sequences \( x(t) \in \mathbb{R} \mapsto y(t) \in \mathbb{R} \) through a \( D \)-dimensional hidden state \( h(t) \in \mathbb{R}^D \). These hidden states evolve over time via parameter matrices \( \mathbf{A} \), \( \mathbf{B} \), and \( \mathbf{C} \), following a linear ordinary differential equation:
\begin{equation}
\begin{aligned}
h^{\prime}(t) =\mathbf{A} h(t)+\mathbf{B} x(t), \quad y(t) =\mathbf{C} h(t),
\end{aligned}
\end{equation}
where parameter \( \mathbf{A} \in \mathbb{R}^{D \times D} \) represents the forgetting gate matrix, \( \mathbf{B} \in \mathbb{R}^{D \times 1} \) denotes the update gate matrix, and \( \mathbf{C} \in \mathbb{R}^{1 \times D} \) serves as the output projection matrix.

To facilitate application in deep learning, SSMs are discretized into discrete-time systems using the zero-order hold technique. The continuous parameters \( \mathbf{A} \) and \( \mathbf{B} \) are transformed into their discrete counterparts \( \mathbf{\overline{A}} \in \mathbb{R}^{D \times D} \) and \( \mathbf{\overline{B}} \in \mathbb{R}^{D \times 1} \), employing a sampling time interval \( \Delta \in \mathbb{R} \):
\begin{equation}
\begin{aligned}
\mathbf{\overline{A}} = \exp(\Delta\mathbf{A}), \quad \mathbf{\overline{B}} = (\Delta\mathbf{A})^{-1}(\exp(\Delta\mathbf{A})-I)\cdot \Delta \mathbf{B}.
\end{aligned}
\end{equation}
Consequently, the discretized SSM can be expressed as follows:
\begin{equation}
\begin{aligned}
h_i = \mathbf{\overline{A}} h_{i-1} +\mathbf{\overline{B}} x_i, \quad y_i = \mathbf{C} h_i,
\end{aligned}
\end{equation}
where \( h_{i-1}, h_i \in \mathbb{R}^{D \times d}\) and \( x_i, y_i \in \mathbb{R}^{1 \times d} \), \(d\) is the dimension of the input sequences.

% \textbf{VideoMamba.} 

\subsection{State Space Prompting}

The backbone of our method is VideoMamba~\cite{li2024videomamba}, the input is a video \( \boldsymbol{V} \in \mathbb{R}^{T \times C \times H \times W} \), where \( T \) represents the number of frames, \( C \) represents the number of channels, and \( H \times W \) is spatial size. Each video frame \( \{\boldsymbol{F}_i\}^{T}_{i=1} \) is split into \( N = \frac{H \times W}{h \times w}\) fixed-size patches of size \( h \times w \) and these patches are flattened into a sequence of vectors \( \boldsymbol{I}_i = \{ \boldsymbol{I}_{ij} \in \mathbb{R}^{C \times h \times w}\}^{N}_{j=1} \), where \( i \) denotes the frame index while \( j \) denotes the patch index. These vectors are then projected into input tokens \( \boldsymbol{x}_i = \{ \boldsymbol{x}_{ij} \}^N_{j=1} \), where \( \boldsymbol{x}_{ij} \in \mathbb{R}^d \), and \( d \) is the hidden dimension of the input sequence. For video classification tasks, the class token \( \boldsymbol{x}_{cls} \) is prepended to the sequence of input tokens, which is \( [\boldsymbol{x}_{cls}, \boldsymbol{x}_1, \boldsymbol{x}_2, \ldots, \boldsymbol{x}_T] \). Then the input tokens are fed into the VideoMamba backbone, which consists of \( L \) layers of Mamba block. The class token \( \boldsymbol{x}_{cls} \) from the last layer is used for classification tasks. The final output of the VideoMamba backbone is obtained by applying a linear classifier head on \( \boldsymbol{x}_{cls} \).

Our SSP consists of two complementary modules: an intra-frame gathering module (IFG) and an inter-frame spreading module (IFS). These modules interact complementarily to aggregate spatial information and spread discriminative long-term context information at the temporal level during fine-tuning. 

% 去除背景色，添加扫描顺序（体现mamba的特点）精简输入输出
% 采样的过程
% 统一一下符号

\subsubsection{Intra-Frame Gathering Module}

The IFG \( \mathcal{P}^s \) processes each frame \( \boldsymbol{F}_i \) to generate intra-frame prompts \( \boldsymbol{p}^s_i \in \mathbb{R}^{N \times d} \), information entropy weights \( \boldsymbol{w}_i \in \mathbb{R}^{1 \times d} \), and spatial variance measurements \( \boldsymbol{v}_i \in \mathbb{R}^{1 \times d} \). The intra-frame prompts are then overlaid to the input tokens to produce spatial prompted tokens \( \boldsymbol{x}^s_i \in \mathbb{R}^{N \times d} \):
\begin{equation}
\boldsymbol{p}^s_i, \boldsymbol{w}_i, \boldsymbol{v}_i = \mathcal{P}^s(\boldsymbol{x}_i), \quad \boldsymbol{x}^s_i = \boldsymbol{x}_i + \boldsymbol{p}^s_i.
\end{equation}

\setlength{\columnsep}{15pt}
\begin{wrapfigure}{r}{0.5\textwidth}
\vspace{-10pt}  % 调整与上方文本的距离
\centering
\includegraphics[width=1\linewidth]{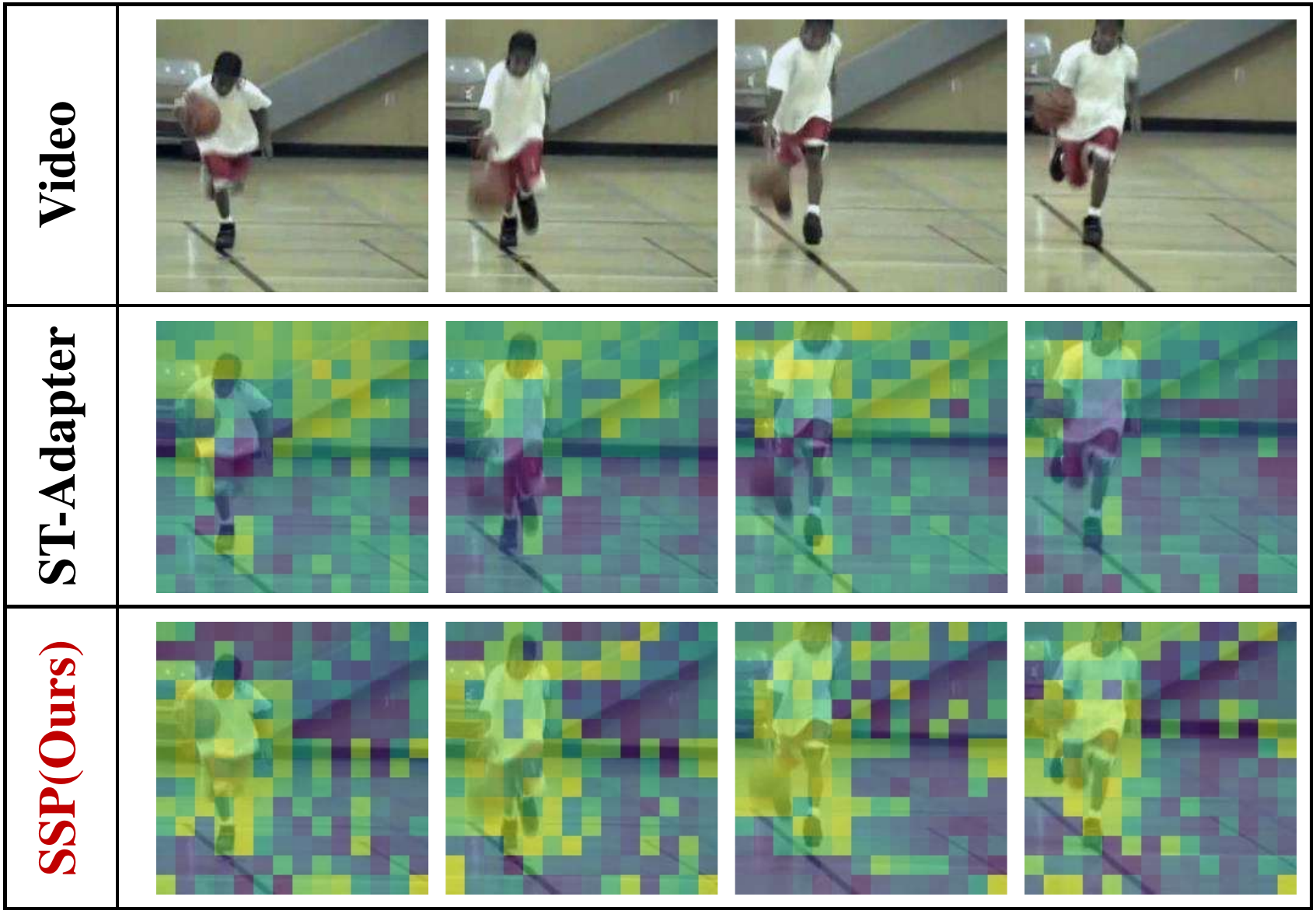}  % 增加图片相对于wrapfigure的宽度
\vspace{-8pt}  % 调整与图像和标题之间的距离
\caption{Visualization of our intra-frame prompts, which capture discriminative local features.}
\label{fig:local_feature}
\vspace{-5pt}  % 调整与下方文本的距离
\end{wrapfigure}

The IFG takes the tokens from each frame \( \boldsymbol{x}_i = \{ \boldsymbol{x}_{ij} \}^N_{j=1} \) as input. These tokens are processed through a downsampling layer \( \mathcal{L}^{down}_1 \) followed by a 2D convolutional layer \( \mathcal{N}^s_1 \) to generate low-rank feature maps \( \boldsymbol{l}_i \in \mathbb{R}^{N \times d^s}\) for each frame, where \( d^s \) represents the internal dimension of the intra-frame gathering module. The low-rank feature maps \( \boldsymbol{l}_i \) are subsequently upsampled via a linear layer \( \mathcal{L}^{up}_1 \) to match the dimensionality of the input tokens, resulting in intra-frame prompts \( \boldsymbol{p}^s_i \). Concurrently, the low-rank feature maps undergo additional 2D convolution and upsampling operations to produce spatial variance \( \boldsymbol{v}_i \). The intra-frame prompts \( \boldsymbol{p}^s_i \) are then fed into an entropy calculation module \( \mathcal{E} \) to calculate information entropy weights \( \boldsymbol{w}_i \):
\begin{equation}
\begin{aligned}
\boldsymbol{l}_i & = \mathcal{N}^s_1(\mathcal{L}^{down}_1(\boldsymbol{x}_i)), \quad &\boldsymbol{p}^s_i = \mathcal{L}^{up}_1(\boldsymbol{l}_i), \\
\boldsymbol{v}_i & = \mathcal{L}^{up}_2(\mathcal{N}^s_2(\boldsymbol{l}_i)), \quad &\boldsymbol{w}_i = \mathcal{E}(\boldsymbol{p}^s_i). \\
\end{aligned}
\end{equation}

The entropy calculation module \( \mathcal{E} \) evaluates the informational significance of intra-frame prompts and generates frame-level weights. This module first transforms the intra-frame prompts \( \boldsymbol{p}^s_i \) into probability distributions, then calculates the information entropy, which is subsequently scaled by a learnable factor \( \alpha \). A small positive constant \(\epsilon\) is incorporated to prevent taking the logarithm of zero:
\begin{equation}
\begin{aligned}
P_i &= \text{softmax}(\boldsymbol{p}^s_i), H_i = -\sum_{d} P_i \cdot \log(P_i + \epsilon), E_i = 1.0 - \frac{H_i}{\max(H_i)}, \\
\boldsymbol{w}_i &= \alpha \cdot \text{softmax}(\bar{E}), \quad \text{where } \bar{E} \text{ is the mean of } E_i \text{ across tokens per frame}.
\end{aligned}
\end{equation}
The intra-frame prompts \( \boldsymbol{p}^s_i \) gather the model's attention to local features during downstream fine-tuning, as shown in Figure \ref{fig:local_feature}. Information entropy weights \( \boldsymbol{w}_i \) adjust the attention given to each frame during inter-frame prompt generation based on the certainty of information distribution, while spatial variance \( \boldsymbol{v}_i \) gates the influence strength from the long-term context information to local features.

\subsubsection{Inter-Frame Spreading Module}

\begin{figure}[H]
    \centering
    \includegraphics[width=1.0\linewidth]{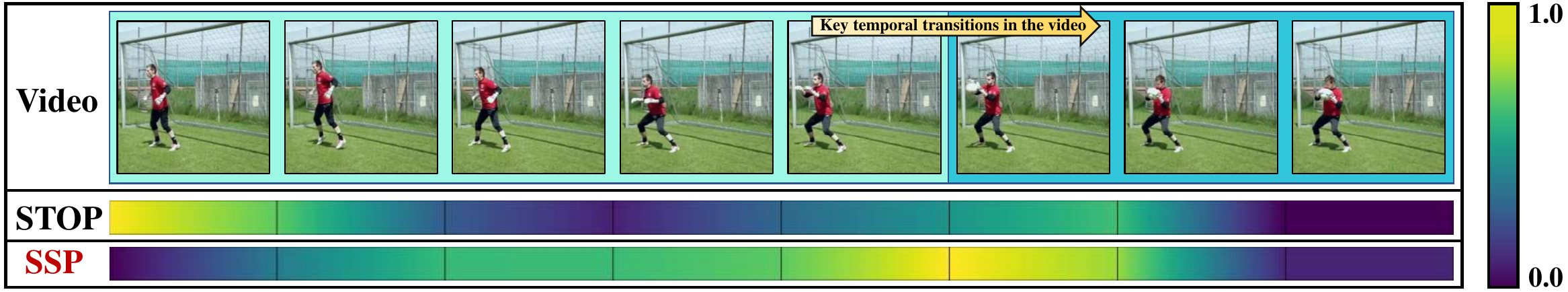}
    \vspace{-12pt}
    \caption{Visualization of our inter-frame prompts. The action "catch" can be parsed into two distinct phases: preparation for the catch and the actual reception of the ball. Our inter-frame prompts effectively locate the key transitional moments between these critical phases.}
    \label{fig:inter_prompt1}
\end{figure}

After processing through the first Mamba layer, the inter-frame spreading module (IFS) activates. This module first samples the last token \( \boldsymbol{s}_i \in \mathbb{R}^{1 \times d} \) from frame \( \boldsymbol{F}_i \) in the Mamba forward scanning sequence. This token is Hadamard multiplied with the information entropy weights \( \boldsymbol{w}_i \) generated by the intra-frame gathering module and fed into the inter-frame spreading module \( \mathcal{P}^t \). The module's output is then Hadamard multiplied with the spatial variance \( \boldsymbol{v}_i \) to produce inter-frame prompts \( \boldsymbol{p}^t_i \in \mathbb{R}^{1 \times d} \):
\begin{equation}
\boldsymbol{s}_i = \boldsymbol{x}_{iN}, \quad \boldsymbol{p}^t_i = \mathcal{P}^t(\boldsymbol{s}_i \odot \boldsymbol{w}_i) \odot \boldsymbol{v}_i.
\label{equation:inter}
\end{equation}
The IFS processes the input through a sequence of operations including a downsampling linear transformation \( \mathcal{L}^{down}_2 \), an attention computation \( \mathcal{A} \), and an upsampling linear transformation \( \mathcal{L}^{up}_3 \). The resulting output is then computed Hadamard product with the spatial variance \( \boldsymbol{v}_i \), and subsequently scaled by a learnable factor \( \beta \) to produce the inter-frame prompts \( \boldsymbol{p}^t_i \). This architecture facilitates the spreading of long-term context information across video frames while preserving local context information:
\begin{equation}
\boldsymbol{p}^t_i = \beta \cdot \mathcal{L}^{up}_3(\mathcal{A}(\mathcal{L}^{down}_2(\boldsymbol{s}_i))) \odot \boldsymbol{v}_i.
\end{equation}
The last token of each frame \( \boldsymbol{s}_i \) aggregates contextual information from both the current frame and preceding frames during forward scanning, as well as subsequent frames during backward scanning. As shown in Figure \ref{fig:inter_prompt1}, the generated inter-frame prompts \( \boldsymbol{p}^t_i \) represent temporal inductive biases that spread the gathered temporal information from all frames to the current frame.

\vspace{-5pt}

\vspace{-5pt}

Subsequently, the inter-frame prompts \( \boldsymbol{p}^t_i \) are concatenated with the spatial prompted tokens \( \boldsymbol{x}^s_i \), along with the class token \( \boldsymbol{x}_{cls} \), are fed into the following \( L - 1 \) Mamba layers \( \{ \mathcal{B}_j \}^
L_{j=2} \) to extract spatio-temporal features. Inspired by VPT-deep~\cite{vpt}, we overlay the intra-frame prompt on the input of each layer and embed the inter-frame prompt in the input of all layers after the first layer. To obtain the final predicted probability distribution \( \boldsymbol{y} \), we apply a linear classification head \( \mathcal{H} \) on \( \boldsymbol{x}_{cls} \).

\subsubsection{Overall Optimization}

As mentioned above, our SSP introduces only a few additional parameters:
\begin{equation}
\boldsymbol{\mathcal{M}} = \{ \mathcal{P}^s, \mathcal{P}^t \}.
\end{equation}
Following prior works, we keep the pre-trained model frozen during training, allowing only the classification head \( \mathcal{H} \) and the newly added modules \( \boldsymbol{\mathcal{M}} \) to be trainable. The optimization objective is defined as follows:
\begin{equation}
\underset{\boldsymbol{\mathcal{M}}, \mathcal{H}}{\arg\min} \ \mathcal{L}_{ce}(\boldsymbol{y}, y_{gt}),
\end{equation}
where \( \mathcal{L}_{ce} \) is the cross-entropy loss, and \( y_{gt} \) is the ground truth video label.

\section{Experiments} \label{sec:experiments}

% 图表把 UCF101 和 SSV2 调换位置

\begin{table}
  \caption{The comparison results on K400 pretrained VideoMamba-S (Parameters 25.42M).}
  \renewcommand{\arraystretch}{1.2}
  \centering
  \small
  \begin{tabularx}{\textwidth}{
      >{\centering\arraybackslash}p{0.3cm} 
    >{\raggedright\arraybackslash}p{2.8cm}@{} 
    >{\raggedright\arraybackslash}p{1.2cm}  | 
    >{\centering\arraybackslash}X |
    >{\centering\arraybackslash}X 
    >{\centering\arraybackslash}X 
    >{\centering\arraybackslash}X 
    >{\centering\arraybackslash}X 
    >{\centering\arraybackslash}X 
    }
  % \begin{tabular}{ll|cc|c|cccc}
    \toprule
      & Method & Venue & Param & HMDB51 & SSV2 & UCF101  & Breakfast \\
    \midrule
    \multirow{9}{*}{\rotatebox{90}{VideoMamba-S}} & 
    Full~\cite{li2024videomamba} & \scriptsize{\textcolor{darkgray}{\textit{\textbf{ECCV'24}}}} 
    & 25.42M & 67.58 & 58.57 & 92.96 & 94.27    \\
    &Adapter~\cite{adapterformer} & \scriptsize{\textcolor{darkgray}{\textit{\textbf{NeurIPS'22}}}} 
    & 2.40M & 73.79 & 36.45 & 94.18 & 84.89    \\
    &ST-Adapter~\cite{st-adapter} & \scriptsize{\textcolor{darkgray}{\textit{\textbf{NeurIPS'22}}}} 
    & 2.69M & 70.52 & 30.94 & 94.87 & 77.60    \\
    &VPT~\cite{vpt} & \scriptsize{\textcolor{darkgray}{\textit{\textbf{ECCV'22}}}} 
    & 1.50M & 72.74 & 30.68 & 95.16 & 81.25    \\
    &VFPT~\cite{vfpt} & \scriptsize{\textcolor{darkgray}{\textit{\textbf{NeurIPS'24}}}} 
    & 1.50M & 72.41 & 30.37 & 95.08 & 79.68    \\
    &SVP~\cite{svp} & \scriptsize{\textcolor{darkgray}{\textit{\textbf{AAAI'25}}}} 
    & 2.76M & 69.93 & 38.01 & 95.58 & 80.72    \\
    &Additional-Scan~\cite{yoshimura2024mambapeft} & \scriptsize{\textcolor{darkgray}{\textit{\textbf{ICLR'25}}}} & 0.66M & 73.20 & 33.71 & 95.63 & 78.64    \\
    &STOP~\cite{STOP} & \scriptsize{\textcolor{darkgray}{\textit{\textbf{CVPR'25}}}} 
     & 1.49M & 70.06 & 21.22  & 93.44 & 65.62    \\
    &SSP(Ours) & \scriptsize{\textcolor{darkgray}{\textit{\textbf{This Paper}}}} & 0.98M & \textbf{74.38} & \textbf{38.68} & \textbf{95.69} & \textbf{85.41}    \\
    \bottomrule
  \end{tabularx}
  \label{table:small-comparison}

\end{table}

\begin{table}
\small
  \caption{The comparison results on CLIP-400M pretrained CLIP-ViT-B/32~\cite{clip} (Parameters 88.00M) and K400 pretrained VideoMamba-M (Parameters 74.00M).}
  \renewcommand{\arraystretch}{1.2}
  \centering
    \begin{tabularx}{\textwidth}{
      >{\centering\arraybackslash}p{0.3cm} 
      >{\raggedright\arraybackslash}p{3cm}@{} 
      >{\raggedright\arraybackslash}p{1.2cm}| 
      >{\centering\arraybackslash}X|
      >{\centering\arraybackslash}X
      >{\centering\arraybackslash}X
      >{\centering\arraybackslash}X
      >{\centering\arraybackslash}X
      >{\centering\arraybackslash}X
    }
    \toprule
    & Method & Venue & Param & HMDB51 & SSV2 & UCF101 & Breakfast \\
    \midrule
    \multirow{3}{*}{\rotatebox{90}{CLIP}} 
     & DGL-Linear~\cite{DGL}           & \scriptsize{\textcolor{darkgray}{\textit{\textbf{AAAI'24}}}}         & 0.83M   & 67.20 & 18.30 & 92.50 & -      \\
    & DGL-Transformer~\cite{DGL}      & \scriptsize{\textcolor{darkgray}{\textit{\textbf{AAAI'24}}}}         & 9.57M   & 69.80 & 18.10 & 93.60 & -      \\
    & STOP~\cite{STOP}                & \scriptsize{\textcolor{darkgray}{\textit{\textbf{CVPR'25}}}}         & 7.53M   & 72.00 & 21.40 & 95.30 & -      \\
    \midrule
    \multirow{9}{*}{\rotatebox{90}{VideoMamba-M}} 
    & Full~\cite{li2024videomamba}    & \scriptsize{\textcolor{darkgray}{\textit{\textbf{ECCV'24}}}}         & 74.00M   & 76.30 & 67.30 & 96.00 & 95.31  \\
    & Adapter~\cite{adapterformer}     & \scriptsize{\textcolor{darkgray}{\textit{\textbf{NeurIPS'22}}}}       & 2.40M    & 73.59 & 46.98 & 96.24 & 89.06  \\
    & ST-Adapter~\cite{st-adapter}   & \scriptsize{\textcolor{darkgray}{\textit{\textbf{NeurIPS'22}}}}       & 2.69M   & 71.96 & 31.91 & 94.81 & 67.70  \\
    & VPT~\cite{vpt}                   & \scriptsize{\textcolor{darkgray}{\textit{\textbf{ECCV'22}}}}         & 1.50M   & 73.59 & 40.58 & 95.45 & 81.77  \\
    & VFPT~\cite{vfpt}                 & \scriptsize{\textcolor{darkgray}{\textit{\textbf{NeurIPS'24}}}}       & 1.50M   & 71.89 & 39.68 & 95.71 & 86.97  \\
    & SVP~\cite{svp}                   & \scriptsize{\textcolor{darkgray}{\textit{\textbf{AAAI'25}}}}         & 2.76M   & 73.66 & 49.08 & 96.77 & 90.10  \\
    & Additional-Scan~\cite{yoshimura2024mambapeft} & \scriptsize{\textcolor{darkgray}{\textit{\textbf{ICLR'25}}}} & 1.33M   & 73.52 & 44.65     & 96.32 & 86.97  \\
    & STOP~\cite{STOP}                 & \scriptsize{\textcolor{darkgray}{\textit{\textbf{CVPR'25}}}}         & 1.49M   & 71.96 & 23.44     & 94.60 & 71.35  \\
    & SSP(Ours)                             & \scriptsize{\textcolor{darkgray}{\textit{\textbf{This Paper}}}}       & 2.41M   & \textbf{76.66} & \textbf{53.72} & \textbf{97.03} & \textbf{93.23} \\
    \bottomrule
  \end{tabularx}
  \label{table:comparison}
\end{table}

\subsection{Datasets}

\textbf{HMDB51}~\cite{HMDB51} contains 6849 clips across 51 action categories, with an average duration of 3.15 seconds and 91.49 frames per video. It was collected from various sources, mostly from movies, and a small proportion from public databases such as the Prelinger archive, YouTube and Google videos, featuring diverse real-world actions with variations in background and camera angles.

\textbf{UCF101}~\cite{UCF101} is an action recognition data set of realistic action videos, collected from YouTube, having 13320 video clips across 101 action categories. The average video length is 7.21 seconds with 186.5 frames per video.

\textbf{Something-Something V2} (SSV2)~\cite{SSV2} is a large-scale dataset for action recognition, containing 220,847 videos across 174 action categories. The dataset is designed to capture fine-grained actions and interactions between objects, with an average video length of 3.82 seconds and 45.84 frames.

\textbf{Breakfast}~\cite{Breakfast} is a long video understanding dataset containing 1989 video clips divided into 10 categories related to breakfast preparation. The dataset has an average video length of 137.53 seconds, with an average of 2062.89 frames per video.

\subsection{Comparison Methods}

We compare our SSP with both adapter-based and prompt-based parameter-efficient finetuning methods. We also report the fully tuning results as a baseline, i.e., VideoMamba~\cite{li2024videomamba}. For adapter-based methods, we compare with the following methods: Adapter~\cite{adapterformer} and ST-Adapter~\cite{st-adapter}. For prompt-based methods, we compare with the following methods: DGL-Linear~\cite{DGL}, DGL-Transformer~\cite{DGL}, VPT~\cite{vpt}, VFPT~\cite{vfpt}, SVP~\cite{svp}, and STOP~\cite{STOP}.

\subsection{Implementation Details} \label{sec:implementation}

Following~\cite{li2024videomamba}, all video frames are resized to \( 224 \times 224 \) and split into \( 14 \times 14 \) patches. For HMDB51, UCF101 and SSV2 datasets, each video is uniformly sampled to 8 frames, while for Breakfast, we sample 32 frames. We set the learning rates to 3e-3, 5e-3, 2e-4, and 1e-3 for HMDB51, UCF101, SSV2, and Breakfast respectively. Meanwhile, we set the batch size to 32 for HMDB51 and UCF101, 64 for Breakfast and 512 for SSV2. All the dataset splits are consistent with the official annotation files. The model is fine-tuned with the AdamW optimizer on 4 NVIDIA 4090-24G GPUs, with a cosine decay scheduler. Additionally, we adopt a warm-up strategy within the first 5 training epochs.

\subsection{Comparison with State-of-the-arts}

We evaluated our SSP method on four popular video datasets: HMDB51, UCF101, SSV2, and Breakfast. For fair comparison, we used CLIP-400M pretrained CLIP-ViT-B/32 as the backbone for ViT-oriented methods, while for methods fine-tuned on the Mamba architecture, we employed Kinetics-400~\cite{kinetics} pretrained VideoMamba-M and VideoMamba-S as backbones. 
Our method achieves state-of-the-art performance across all datasets, as shown in Table \ref{table:comparison}.
\setlength{\columnsep}{15pt}
\begin{wrapfigure}{r}{0.5\textwidth}
\vspace{-10pt}  % 调整与上方文本的距离
\centering
\includegraphics[width=1\linewidth]{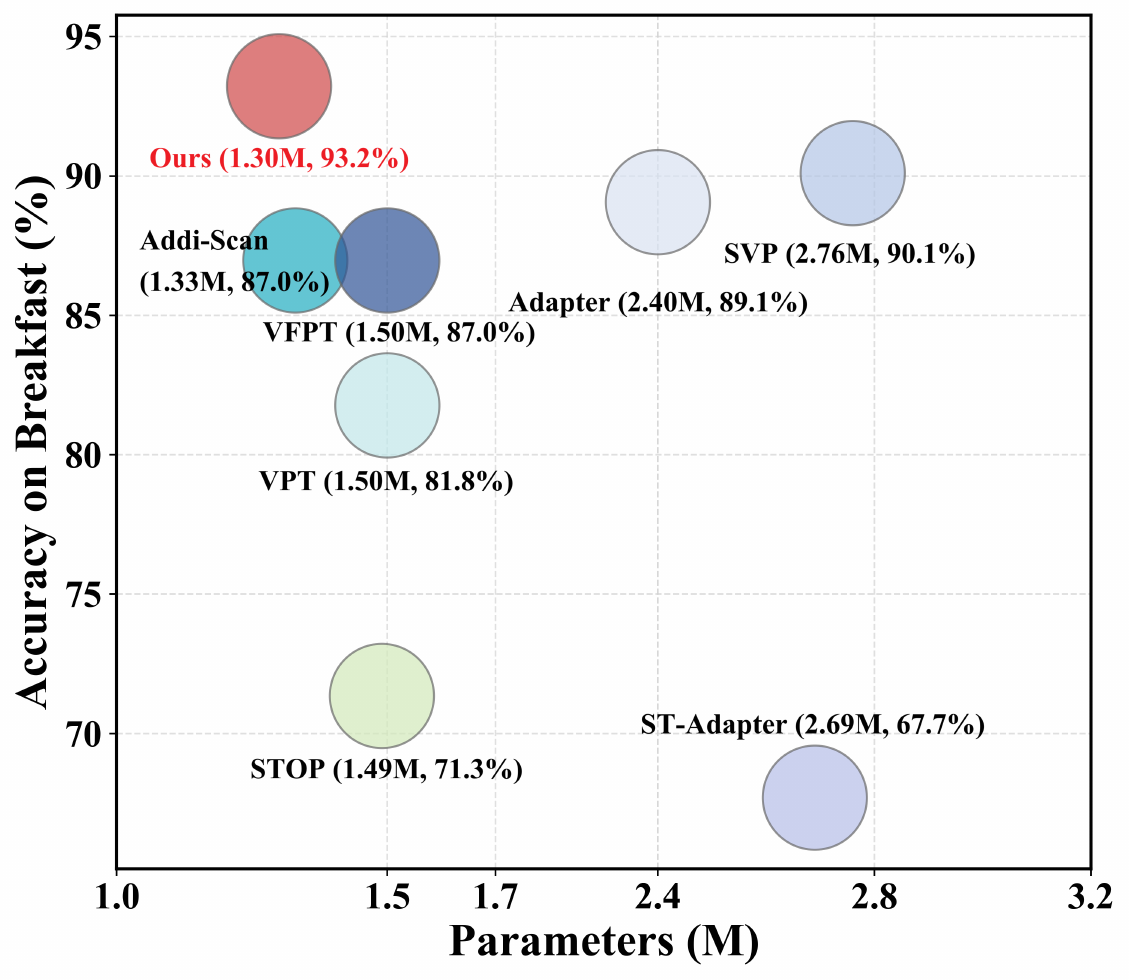}  % 增加图片相对于wrapfigure的宽度
\vspace{-12pt}  % 调整与图像和标题之间的距离
\caption{Comparison on Breakfast. SSP outperforms existing methods while tuning minimal parameters.}
\label{fig:2d-compare}
\end{wrapfigure}
When fine-tuned on VideoMamba-M, our approach attains top-1 accuracies of 76.66\%, 53.72\%, 97.03\%, and 93.23\% on HMDB51, Something-Something V2, UCF101, and Breakfast respectively, representing improvements of \textbf{3.00\%}, \textbf{4.64\%}, \textbf{0.26\%}, and \textbf{3.13\%} over existing parameter-efficient video fine-tuning methods. As demonstrated in Table \ref{table:small-comparison}, even with the parameter constraints of smaller-scale models, our method still delivers superior results when fine-tuned on VideoMamba-S. This effectiveness stems from our tailored design for the VideoMamba architecture, which enables efficient gathering and spreading of discriminative spatio-temporal information within state space models, through the complementary IFG and IFS modules. Notably, our approach demonstrates particularly significant improvements on the challenging large-scale Something-Something V2 dataset and the long-video Breakfast dataset, as shown in Figure \ref{fig:2d-compare}, our method exceeds others when tuning only 1.30M parameters (using only one IFS module as shown in Figure \ref{fig:abla_inter_num}). This is because our SSP can gather local information on key regions in complex video data and spread critical global information in long contexts.

\subsection{Ablation}

\subsubsection{Influence of Different Components}

To verify the effectiveness of the intra-frame gathering (IFG) module and inter-frame spreading (IFS) module, we conducted ablation experiments on three datasets: HMDB51, UCF101, and Breakfast, as shown in Table \ref{table:ablation_combined}. As demonstrated, when neither component is used, SSP achieves the lowest accuracy on all datasets. When the intra-frame gathering module is used alone, the model's performance improves by 9.78\% on average. This is because the intra-frame spatial prompts effectively capture local features and enhance the model's ability to focus on discriminative information within each frame. When the inter-frame spreading module is used alone, the model's performance improves by 9.20\% on average. This can be attributed to that the inter-frame temporal prompts effectively aggregate and spread global contextual information across frames. When both components are used together, the model achieves the best performance by an average improvement of 13.46\% across all datasets, as they complement each other in gathering and spreading both local and global information. When the spatial variance gate or the entropy gate (denoted as \(\boldsymbol{v}_i, \boldsymbol{w}_i\) in Equation \ref{equation:inter}) is removed, the model's performance drops by 2.33\% and 2.27\% respectively on average across all datasets. This indicates that both gates play a crucial role in facilitating the inter-frame prompts to propagate spatial information in a complementary manner based on the key local features of each frame.

\begin{table}[ht]
\small
  \caption{Ablation of different prompting modules and gates.}
  \renewcommand{\arraystretch}{1.2}
  \centering
  \begin{tabularx}{\textwidth}
  {
    % @{}
    >{\centering\arraybackslash}p{0.6cm}
    >{\centering\arraybackslash}p{0.6cm}|
    >{\centering\arraybackslash}p{1.1cm}
    >{\centering\arraybackslash}p{1cm}
    >{\centering\arraybackslash}X||
    >{\centering\arraybackslash}p{0.9cm}
    >{\centering\arraybackslash}p{0.9cm}|
    >{\centering\arraybackslash}p{1.1cm}
    >{\centering\arraybackslash}p{1cm}
    >{\centering\arraybackslash}X
  }
  % {cc|ccc||cc|ccc}
    \toprule
    \multicolumn{5}{c||}{\textbf{Different Prompting Modules}} & \multicolumn{5}{c}{\textbf{Different Gates}} \\
    \hline
    % Intra & Inter & HMDB51 & UCF101 & Breakfast & Entropy & Spatial & HMDB51 & UCF101 & Breakfast \\
    \rowcolor{mygray}
    IFG & IFS & HMDB51 & UCF101 & Breakfast & Entropy & Spatial & HMDB51 & UCF101 & Breakfast \\
    \hline
    - & - & 59.34 & 90.64 & 76.56 & - & - & 74.70 & 96.56 & 89.58 \\
    \ding{52} & - & 74.44 & 96.56 & 84.89 & \ding{52} & - & 75.09 & 96.29 & 88.54 \\
    - & \ding{52} & 72.61 & 96.14 & 85.41 & - & \ding{52} & 74.96 & 96.61 & 88.54 \\
    \rowcolor{mygray}
    \ding{52} & \ding{52} & \textbf{76.66} & \textbf{97.03} & \textbf{93.23} & \ding{52} & \ding{52} & \textbf{76.66} & \textbf{97.03} & \textbf{93.23} \\
    \bottomrule
  \end{tabularx}
  \label{table:ablation_combined}
\end{table}

\subsubsection{The Visualization Results of Update Gate}

\begin{figure}[H]
    \centering
    \includegraphics[width=1.0\linewidth]{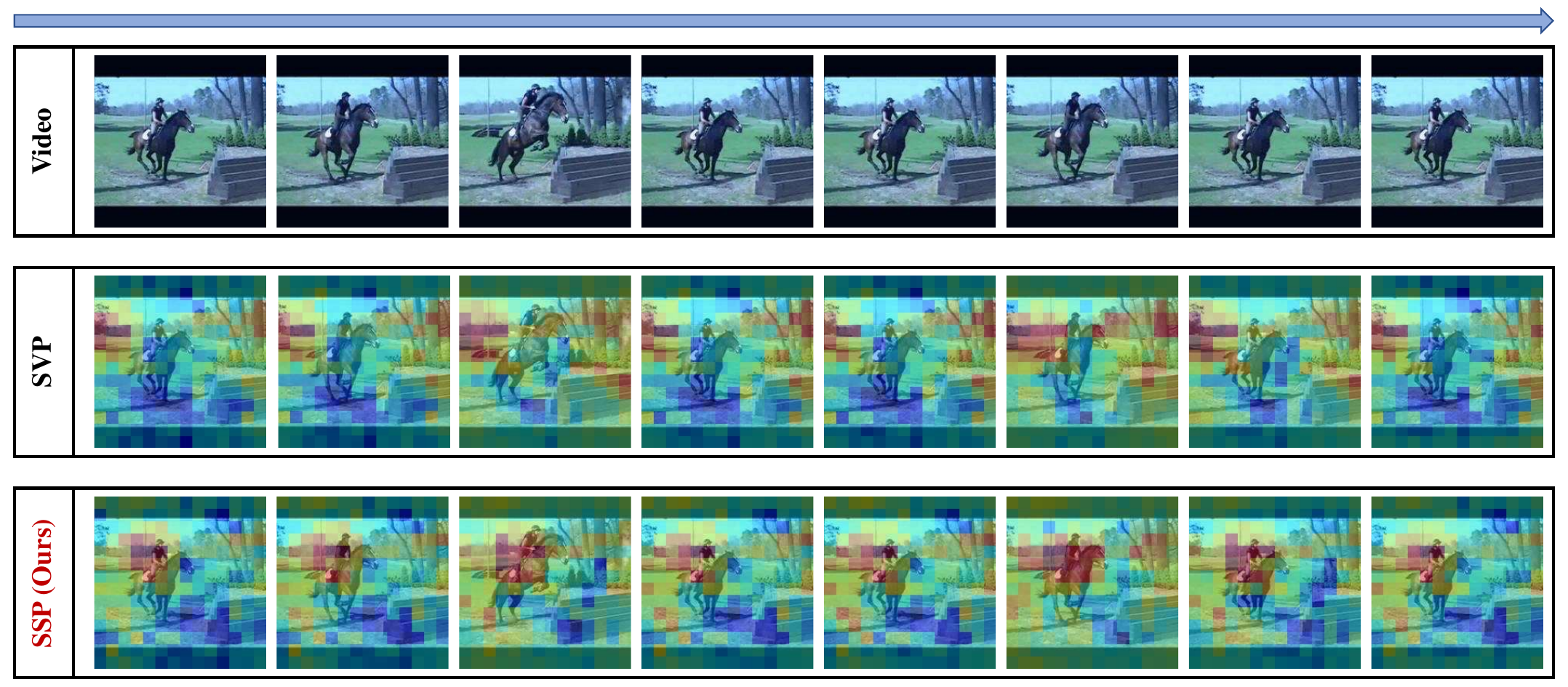}
    \vspace{-12pt}
    \caption{Visualization of the update
 gate. In SSP, the key region is activated effectively.}
    \label{fig:updating gate}
\end{figure}

\vspace{-5pt}

\setlength{\columnsep}{15pt}
\begin{wrapfigure}{l}{0.70\textwidth}  % 减小宽度以减少右侧空白
\vspace{-10pt}  % 调整与上方文本的距离
\centering
\includegraphics[width=1\linewidth]{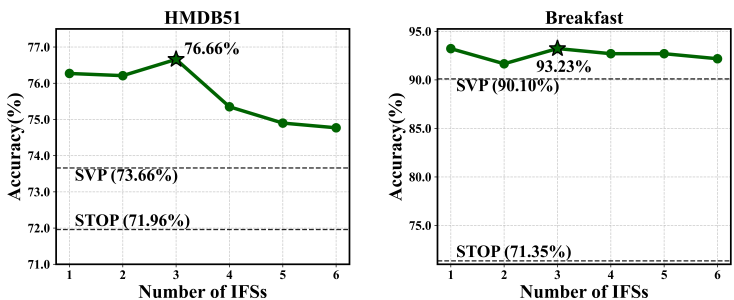}  % 增加图片相对于wrapfigure的宽度
\vspace{-10pt}  % 调整与图像和标题之间的距离
\caption{Ablation of the number of IFSs in our SSP.}
\label{fig:abla_inter_num}
\vspace{-15pt}  % 调整与下方文本的距离
\end{wrapfigure}

To further explore the impact of our intra-frame gathering module and inter-frame spreading module, we visualize the normalized update gate values in VideoMamba on the last layer, which contribute to the final classification results directly. As shown in Figure \ref{fig:updating gate}, existing prompting methods designed for Mamba like SVP can't effectively gather and spread the discriminative spatio-temporal information. This caused the pre-trained VideoMamba model to focus on irrelevant regions in each frame. As a result, the model struggles to accurately understand dynamic key features in the video. In contrast, our SSP method highlight the key regions with dynamic changes in the video, leading to a more accurate understanding of the video content.

\subsubsection{Influence of Hyperparameters} \label{sec:IFS_number}

The number of the inter-frame spreading modules (\( \mathcal{P}^t \)) is one of important hyperparameters in our method. To assess its impact, we conduct extensive ablation experiments. As shown in Figure \ref{fig:abla_inter_num}, the model's performance initially improves but then fluctuates as the number of inter-frame spreading modules increases. 

To balance performance and tunable parameters, we set the number of inter-frame spreading modules to 3 when using VideoMamba-M as backbone, and when using VideoMamba-S as backbone, we set the number to 1.

\section{Conclusion}

In this paper, we propose a novel State Space Prompting (SSP) approach for efficient adaptation of pre-trained state space models to video understanding tasks. SSP combines complementary intra-frame gathering module and inter-frame spreading module to aggregate key spatial information within frames and spread discriminative temporal information between frames, enabling effective propagation of crucial spatio-temporal features in state space models. By employing the two modules, we can adaptively balance and compress discriminative information in videos. The effectiveness of our proposed SSP has been validated on four video benchmarks compared to other parameter-efficient fine-tuning methods.

\section{Acknowledgements}

This work was supported by the National Natural Science Foundation of China (62376011) and the National Key R\&D Program of China (2024YFA1410000).

%\begin{table}[H]
  %\caption{CLS Token Position}
  %\label{sample-table}
  %\renewcommand{\arraystretch}{1.5}
  %\centering
  %\begin{tabular}{l|cc}
    %\toprule
    %Method & HMDB51 & UCF101 \\
    %\midrule
    %CLS-mid & 75.55 & 96.59    \\
    %CLS-back & 75.29 & 96.64    \\
    %CLS-pre & \textbf{76.20} & \textbf{96.98}    \\
    %\bottomrule
  %\end{tabular}
%\end{table}

% \section{References}

\bibliographystyle{unsrt}
\bibliography{reference}

%\subsection{Math}
%Note that display math in bare TeX commands will not create correct line numbers for submission. Please use LaTeX (or AMSTeX) commands for unnumbered display math. (You really shouldn't be using \$\$ anyway; see \url{https://tex.stackexchange.com/questions/503/why-is-preferable-to} and \url{https://tex.stackexchange.com/questions/40492/what-are-the-differences-between-align-equation-and-displaymath} for more information.)

%\subsection{Final instructions}

%Do not change any aspects of the formatting parameters in the style files.  In
%particular, do not modify the width or length of the rectangle the text should
%fit into, and do not change font sizes (except perhaps in the
%\textbf{References} section; see below). Please note that pages should be
%numbered.

%{
%\small

%[1] Alexander, J.A.\ \& Mozer, M.C.\ (1995) Template-based algorithms for
%connectionist rule extraction. In G.\ Tesauro, D.S.\ Touretzky and T.K.\ Leen
%(eds.), {\it Advances in Neural Information Processing Systems 7},
%pp.\ 609--616. Cambridge, MA: MIT Press.

%[2] Bower, J.M.\ \& Beeman, D.\ (1995) {\it The Book of GENESIS: Exploring
%  Realistic Neural Models with the GEneral NEural SImulation System.}  New York:
%TELOS/Springer--Verlag.

%[3] Hasselmo, M.E., Schnell, E.\ \& Barkai, E.\ (1995) Dynamics of learning and
%recall at excitatory recurrent synapses and cholinergic modulation in rat
%hippocampal region CA3. {\it Journal of Neuroscience} {\bf 15}(7):5249-5262.
%}

% %%%%%%%%%%%%%%%%%%%%%%%%%%%%%%%%%%%%%%%%%%%%%%%%%%%%%%%%%%%%

% \appendix

\input{nips_checklist}

\clearpage

\appendix

\section{Discussion and Analysis}

In this section, we discuss and analyze the effectiveness of our SSP method for parameter-efficient fine-tuning the VideoMamba architecture in downstream task adaptation.

For each Mamba layer's input, we denote the \(i\)-th input token along the scanning order as \(\boldsymbol{x}_i\). The hidden state computation for the subsequent Mamba layer is formulated as:
\begin{equation}
\boldsymbol{h}_{i} = \mathbf{\overline{A}}_i \boldsymbol{h}_{i-1} + \mathbf{\overline{B}}_i \boldsymbol{x}^s_i, \quad \boldsymbol{x}^s_i = \boldsymbol{x}_i + \boldsymbol{p}^s_i,
\label{equation:state}
\end{equation}
where \( \boldsymbol{h}_{i} \) is the \( i \)-th compressed hidden state token of the Mamba layer. In the Mamba architecture, each layer's parameter matrices \(\mathbf{B} \in \mathbb{R}^{\mathrm{D} \times 1}\), \(\mathbf{C} \in \mathbb{R}^{1 \times \mathrm{D}}\), and \(\Delta \in \mathbb{R}\) are derived from the input token \(\boldsymbol{x}^s_i\) through functions \(\mathcal{S}_B\), \(\mathcal{S}_C\), and \(\mathcal{S}_\Delta\), respectively:
\begin{equation}
\mathbf{\overline{A}}_i = \exp(\Delta\mathbf{A}_i), \quad
\mathbf{B}_i = \mathcal{S}_B(\boldsymbol{x}^s_i), \quad \mathbf{C}_i = \mathcal{S}_C(\boldsymbol{x}^s_i), \quad \Delta_i = \mathcal{S}_\Delta(\boldsymbol{x}^s_i).
\end{equation}
In our proposed SSP method, the intra-frame prompt \(\boldsymbol{p}^s_i\) enables direct fine-tuning of the parameter matrices generated by each token:
\begin{equation}
\overline{\mathbf{A}}_i^{p} = \exp(\mathcal{S}_\Delta(\boldsymbol{x}_i + \boldsymbol{p}^s_i) \widetilde{\odot} \mathbf{A}), \quad
\overline{\mathbf{B}}_i^{p} = \mathcal{S}_{\Delta}(\boldsymbol{x}_i + \boldsymbol{p}^s_i)\mathcal{S}_B(\boldsymbol{x}_i + \boldsymbol{p}^s_i),
\label{equation:parameterized matrix}
\end{equation}
where \(\overline{\mathbf{A}}_i^p \in \mathbb{R}^{\mathrm{D} \times \mathrm{D}}\) and \(\mathbf{\overline{B}}_i^p \in \mathbb{R}^{\mathrm{D} \times 1}\) represent the forget and update gates directly controlled by the intra-frame prompt \(\boldsymbol{p}^s_i\), facilitating the extraction of locally discriminative information, thereby gathering the spatial information effectively when fine-tuning on downstream tasks.

For the inter-frame prompt \(\boldsymbol{p}^t_j\) of the \(j\)-th frame, this prompt vector aggregates global spatio-temporal information and is gated through Hadamard multiplication with the spatial variance \(\boldsymbol{v}_j\). It is inserted between the \(j\)-th and \((j+1)\)-th frames, directly influencing the hidden state at that position:
\begin{equation}
\boldsymbol{h}_{j+1, 0} = \mathbf{\overline{A}}_{j+1, 0}^p \boldsymbol{h}_{jN} + \mathbf{\overline{B}}_{j+1,0}^p \boldsymbol{p}^t_j,
\end{equation}
where \(N\) denotes the number of tokens per frame. Tokens following the \(j\)-th frame can access global discriminative information through the hidden state \(\boldsymbol{h}_{j+1, 0}\) of inter-frame prompt \(\boldsymbol{p}^t_j\), overcoming the sequential spatio-temporal information transfer limitation of the original VideoMamba model and achieving efficient gathering and spreading of spatio-temporal information.

Next, we further analyze the impact of inter-frame prompts on long-range spatio-temporal information transmission. By expanding Equation \ref{equation:state}, we obtain:
\begin{equation}
\begin{aligned}
\boldsymbol{h}_{j} = \mathbf{\overline{A}}_j^p \boldsymbol{h}_{j-1} + \mathbf{\overline{B}}_j^p \boldsymbol{x}^s_j & = \mathbf{\overline{A}}_j^p \left(\mathbf{\overline{A}}_{j-1}^p \boldsymbol{h}_{j-2} + \mathbf{\overline{B}}_{j-1}^p \boldsymbol{x}^s_{j-1}\right) + \mathbf{\overline{B}}_j^p \boldsymbol{x}^s_j \\
& = \prod_{k=i+1}^{j} \mathbf{\overline{A}}_{k}^p \boldsymbol{h}_{i} + \sum_{t=i+1}^{j} \left( \prod_{k=t+1}^{j} \mathbf{\overline{A}}_{k}^p \right) \mathbf{\overline{B}}_{t}^p \boldsymbol{x}^s_t,
\end{aligned}
\end{equation}
where \(i < j\). The coefficient of the first term in the above equation, \(\prod_{k=i+1}^{j} \mathbf{\overline{A}}_{k}^p\), represents the influence strength from the \(i\)-th hidden state token to the \(j\)-th hidden state token along the scanning sequence. We denote this as the transmission matrix \(\mathbf{T}_{i \rightarrow j}\):
\begin{equation}
\mathbf{T}_{i \rightarrow j} = \prod_{k=i+1}^{j} \mathbf{\overline{A}}_{k}^p = \exp\left(\sum_{k=i+1}^{j} \Delta_k \widetilde{\odot} \mathbf{A}\right),
\end{equation}
where \(\mathbf{A}\) is a negative matrix. Examining the equation above, we observe that the mutual influence strength between tokens at different positions in the sequence exponentially decays to zero as the distance \(j - i\) increases. Without the insertion of inter-frame prompts, the maximum information transmission path length in the sequence equals the sequence length \(\mathcal{O}(TN)\), where \(T\) is the number of video frames and \(N\) is the number of tokens per frame. However, after inserting inter-frame prompts, information from tokens at different positions can be transmitted through the inter-frame prompts inserted after each frame, reducing the maximum information transmission path length to \(\mathcal{O}(N)\). This significant reduction in information transmission path length facilitates the aggregation and propagation of global spatio-temporal information.

\clearpage

\section{Visualization of More Cases}

\subsection{Update Gate Value Visualization}

\begin{figure}[H]
    \centering
    \includegraphics[width=1.0\linewidth]{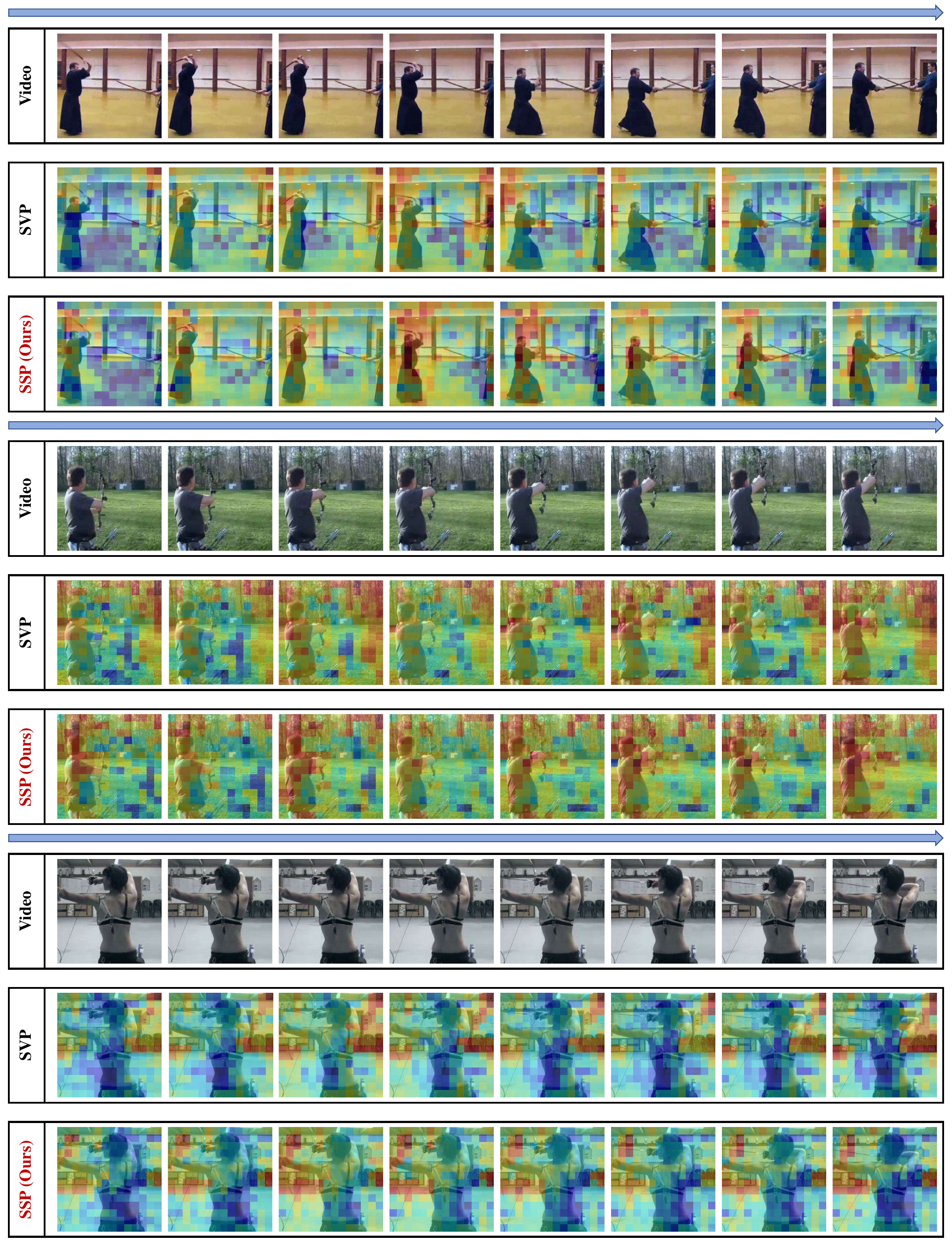}
    \vspace{-12pt}
    \caption{Visualization of the values of the update gate over the last layer in VideoMamba.}
    \label{fig:updating gate appendix}
\end{figure}

\vspace{-5pt}

The update gate value of each token, denoted as \( \overline{\mathbf{B}}_i^{p} \) in Equation \ref{equation:parameterized matrix}, represents the degree of influence each token exerts on the hidden states of VideoMamba. By visualizing the update gate values across tokens, we can observe which specific tokens receive greater attention from the VideoMamba architecture during processing.

As illustrated in Figure \ref{fig:updating gate appendix}, existing prompting approaches for Mamba architecture (e.g., SVP) employ static prompts for each frame. This limitation prevents the pre-trained VideoMamba model from effectively integrating and modeling global video contextual information, thereby hindering accurate interpretation of human actions within the video sequence. In contrast, our SSP method, through the complementary application of intra-frame gathering and inter-frame spreading modules, dynamically emphasizes regions exhibiting significant temporal variations in the video. This approach enables more precise comprehension of video content by capturing the most relevant spatio-temporal information.

\subsection{Intra-Frame Prompts Visualization}

In our SSP method, the intra-frame prompts, which are overlaid on the input tokens fed into each Mamba layer, are employed to capture the discriminative local features and gather the spatial information of each frame. To visualize the intra-frame prompts, we plot the values of intra-frame prompts of all frames in a video over the last Mamba layer as heatmaps, and overlay them on the original video frames.

\begin{figure}[H]
    \centering
    \includegraphics[width=1.0\linewidth]{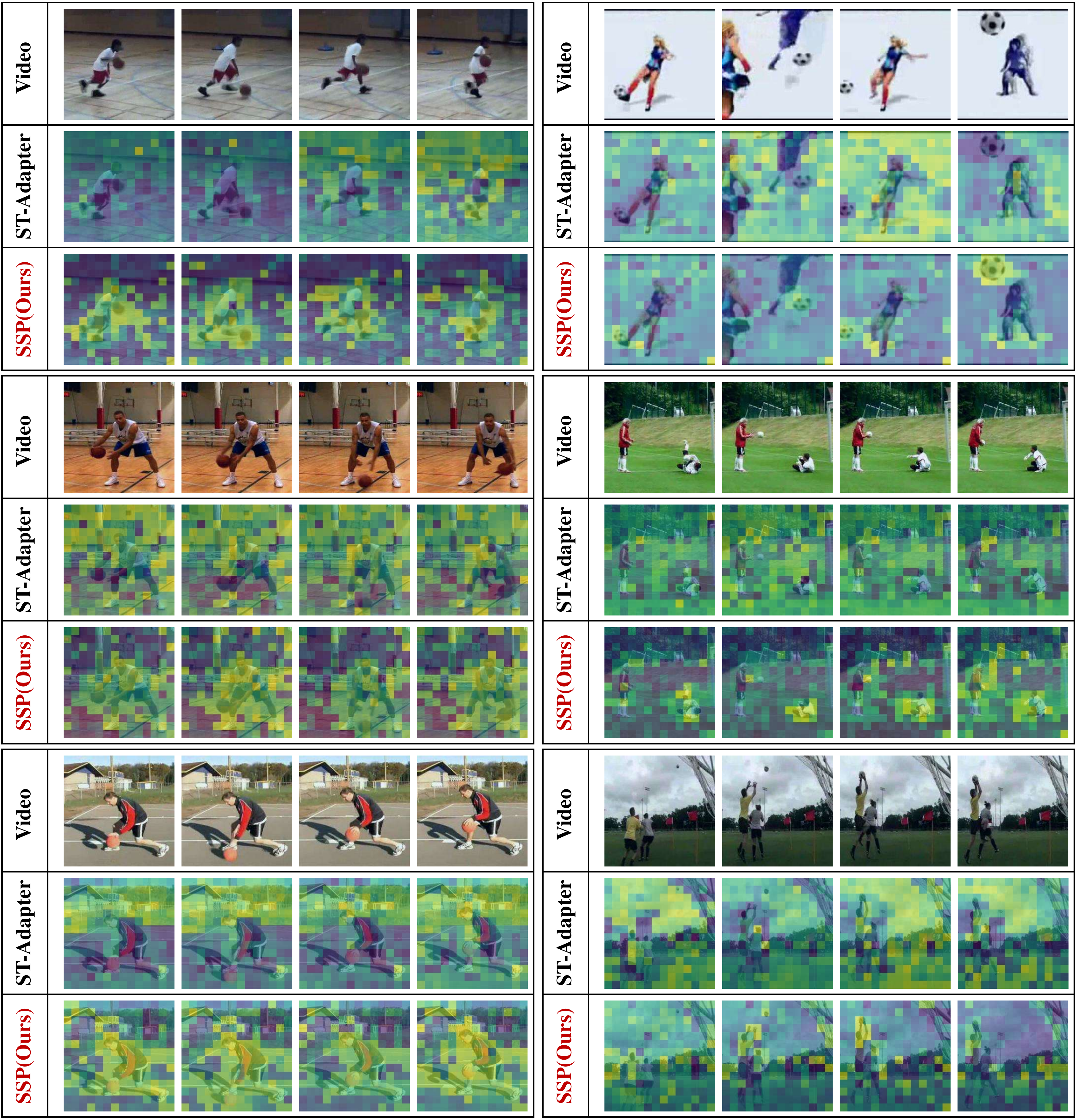}
    \vspace{-12pt}
    \caption{Visualization of our intra-frame prompts
over the last layer in VideoMamba.}
    \label{fig:local features appendix}
\end{figure}

\vspace{-5pt}

As shown in Figure \ref{fig:local features appendix}, when existing parameter-efficient fine-tuning methods (e.g., ST-Adapter) are applied to VideoMamba, they fail to effectively capture the local feature maps of individual frames, preventing the model from attending to local features that undergo temporal variations. In contrast, our approach, through efficient spreading of global temporal information, enables intra-frame prompts to more accurately capture discriminative local features, thereby achieving effective gathering of spatially relevant information.

\subsection{Inter-Frame Prompts Visualization}

The inter-frame prompts in our SSP method facilitate the propagation of temporal information across sequential frames. Through visualization of these inter-frame prompts, we can identify which specific frames receive heightened attention from the model during global context integration, providing insights into the temporal dynamics of information processing within the architecture.

\begin{figure}[H]
    \centering
    \includegraphics[width=1.0\linewidth]{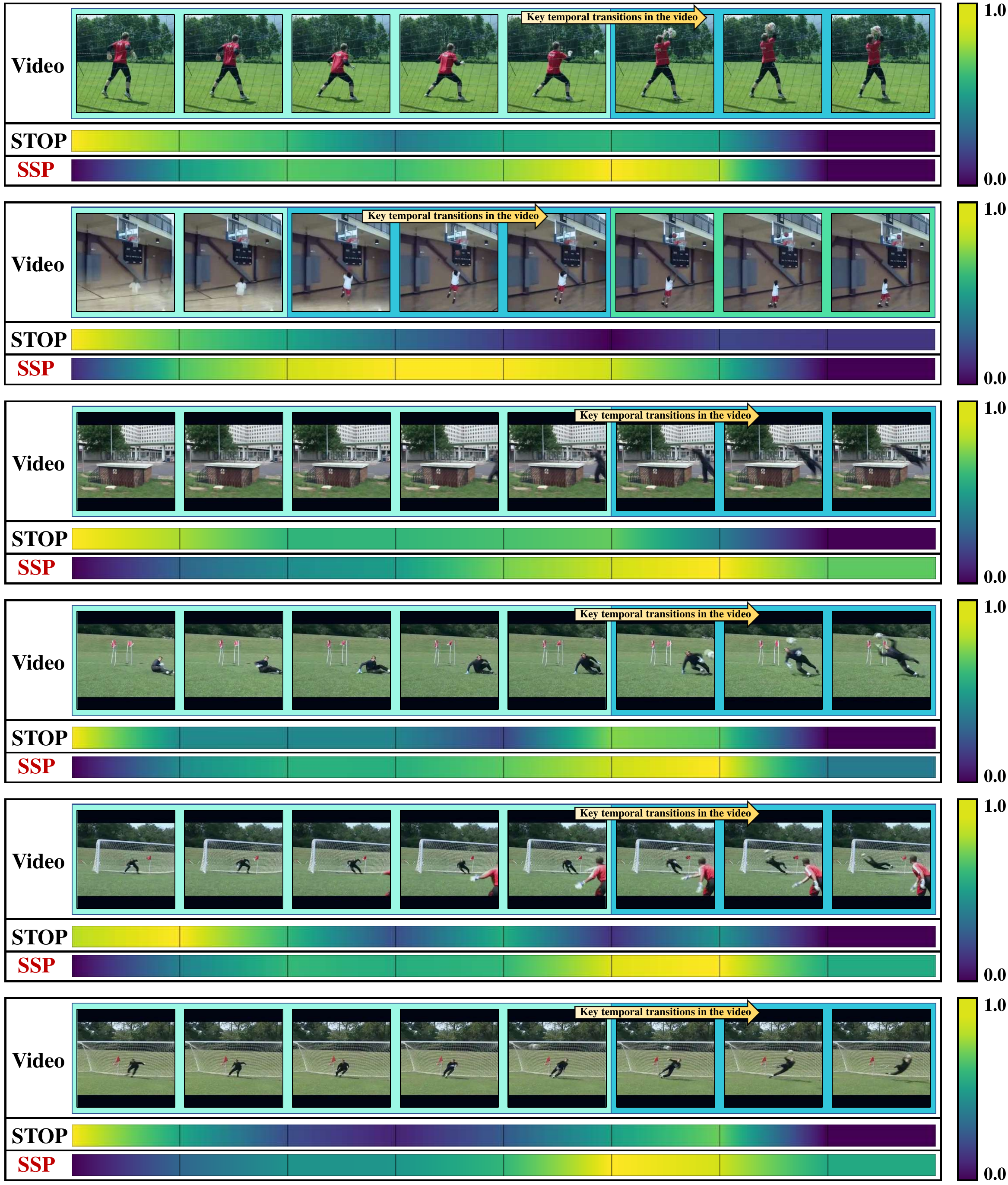}
    \vspace{-12pt}
    \caption{Visualization of our inter-frame prompts over the last layer in VideoMamba.}
    \label{fig:inters appendix}
\end{figure}

\vspace{-5pt}

As demonstrated in Figure \ref{fig:inters appendix}, compared to existing video prompting methods (e.g., STOP), the inter-frame prompts of our SSP method effectively identify key frames exhibiting temporal variations within the video sequence. This enhanced capability stems from our approach being specifically engineered for the Mamba architecture, enabling efficient refinement and propagation of temporal information from compressed hidden states.

\section{Discussion on Different Method Designs}

\begin{wraptable}{r}{0.55\linewidth}
\vspace{-15pt}
\small
\centering
% \setlength{\abovecaptionskip}{0cm}
% \setlength{\belowcaptionskip}{-0.2cm}
% \hspace{-0.6em}
\caption{Ablation of Different Sample Methods.}
\renewcommand{\arraystretch}{1.2}
\vspace{0.5em}
\label{table:swin}
\begin{tabular}{c|ccc} 
\toprule
\rowcolor{mygray}
\footnotesize  Method & HMDB51 & UCF101 & Breakfast \\ 
\hline
\footnotesize  Middle & 74.44 & 95.87 & 91.14 \\
\footnotesize  Bidirection & 75.81  & 96.00 & 89.06   \\
\footnotesize  Bi-Independent & 75.16  & 96.06 & 90.10   \\
\hline
\rowcolor{mygray}
\footnotesize  SSP(Ours) & \textbf{76.66} & \textbf{97.03} & \textbf{93.23} \\
\bottomrule
\end{tabular}
\label{table:sampling}
\vspace{-0.5em}
\end{wraptable}

\begin{figure}[H]
    \centering
    \includegraphics[width=1.0\linewidth]{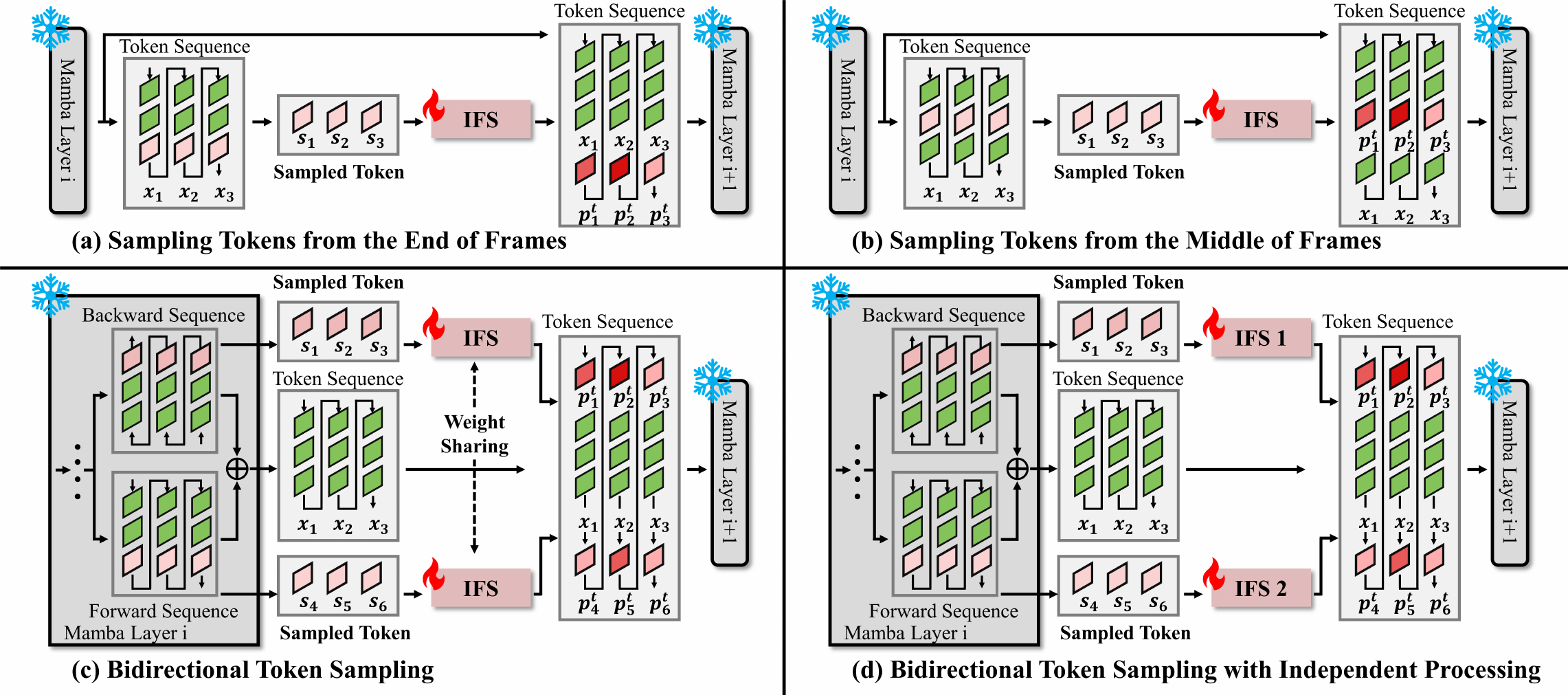}
    \vspace{-12pt}
    \caption{Different methods for inter-frame prompting. For simplicity, irrelevant computational processes in the Mamba block and intra-frame gathering modules have been omitted.}
    \label{fig:sample}
\end{figure}

In our methodology design, we generate inter-frame prompts by sampling the last token from each frame in the forward scanning sequence, as illustrated in Figure \ref{fig:sample}(a). This section explores how different sampling strategies affect the generation of inter-frame prompts, with supplementary experimental results presented in Table \ref{table:sampling}. We discovered that sampling tokens from the middle of each frame (Figure \ref{fig:sample}(b)) leads to an average accuracy decrease of 1.82\%. We attribute this decline to the fragmentation of semantic information within each frame during prompt generation. When sampling the last tokens separately from the forward and backward scanning sequences to generate prompts (Figure \ref{fig:sample}(c)), the accuracy decreases by an average of 2.01\%. We posit that sampling before the superposition of bidirectional sequences causes the inter-frame prompts to overlook complementary information from both directions, resulting in a separation of forward and backward contextual cues. Additionally, employing independent inter-frame spreading modules to generate prompts for forward and backward directions (Figure \ref{fig:sample}(d)) still results in an accuracy drop of 1.86\%. Although this approach introduces more learnable parameters to separately model contextual relationships in forward and backward scanning, it fails to address the fundamental issue of separated forward and backward contextual cues.

\vspace{-5pt}

\section{More Ablation Studies on Hyperparameters} \label{sec:dimension}

The internal dimensions of our intra-frame gathering module (IFG) and inter-frame spreading module (IFS) are also hyper-parameters of our method. To balance efficiency and effectiveness, we set the internal dimension of IFG to 384 and the internal dimension of IFS to 256. We conducted extensive experiments on HMDB51 and Breakfast datasets to investigate the impact of different dimension settings on fine-tuning performance.

\begin{figure}[H]
    \centering
    \includegraphics[width=1.0\linewidth]{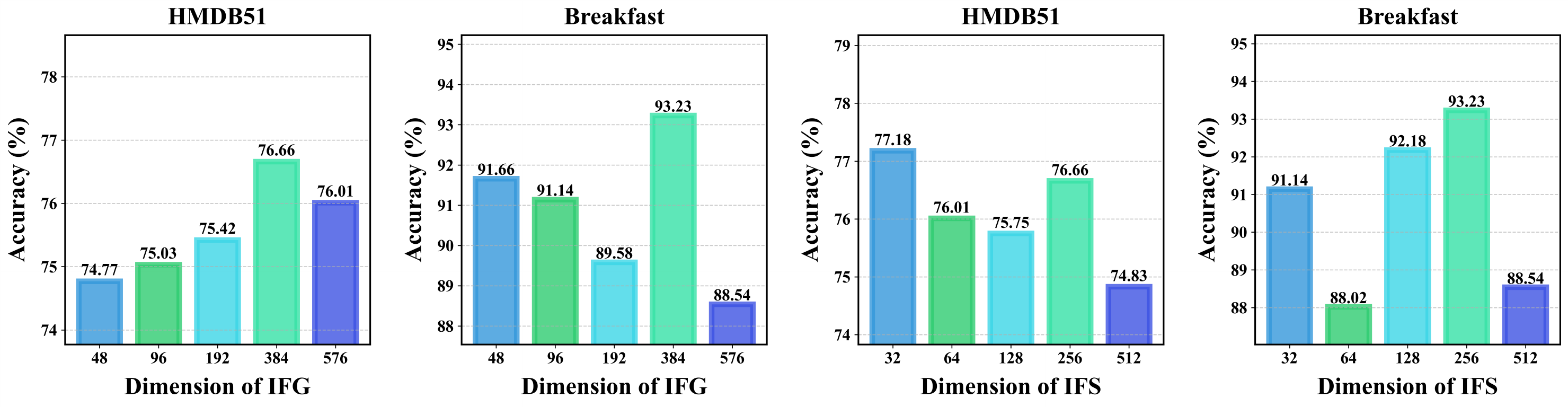}
    \vspace{-12pt}
    \caption{Ablation on the Internal Dimensions of IFG and IFS.}
    \label{fig:dimension}
\end{figure}

As shown in Figure \ref{fig:dimension}, naively increasing the internal dimension of the IFG module does not improve model performance. When the dimension is too large, model performance actually decreases due to increased optimization difficulty. Regarding the internal dimension of the IFS module, the long-video Breakfast dataset is more sensitive to this setting. From the perspective of reducing training time and parameter costs, setting the internal dimension of the IFS to 32 is also an acceptable option, which demonstrates the robustness of our method to the choices of hyperparameters.

\vspace{-5pt}

\section{More Clarifications}

We first clarify the role of spatial variance in our approach. Given that different video frames exhibit varying degrees of association with global information, we learn the spatial variance \( \boldsymbol{v}_i \) through the spatial features aggregated by IFG. The spatial variance gates the weight of each inter-frame prompt, thereby controlling the degree of influence each video frame has on global information and achieving more refined temporal information propagation.

To further elaborate on the spreading mechanism, we describe how IFG and IFS work collaboratively. The IFG module aggregates spatial information from each frame into intra-frame prompts and superimposes the intra-frame prompts onto tokens. While the IFS module performs global attention computation through sampled tokens, enabling global interaction of the aggregated spatial information to generate inter-frame prompts. The inter-frame prompt corresponding to each frame thus contains spatial information from other frames, thereby propagating the aggregated local spatial information in a temporal manner.

\section{Asset License and Consent} \label{sec:asset}

\href{https://github.com/knightyxp/DGL/blob/main/LICENSE}{DGL}, \href{https://github.com/KMnP/vpt/blob/main/LICENSE}{VPT} and \href{https://github.com/runtsang/VFPT/blob/master/LICENSE}{VFPT} are licensed under CC-BY-NC 4.0. \href{https://github.com/ShoufaChen/AdaptFormer/blob/main/LICENSE}{Adapter}, \href{https://github.com/linziyi96/st-adapter/blob/main/LICENSE}{ST-Adapter}, and \href{https://github.com/openai/CLIP/blob/main/LICENSE}{CLIP-ViT} are licensed under MIT. \href{https://github.com/zhoujiahuan1991/CVPR2025-STOP/blob/main/LICENSE}{STOP}, \href{https://github.com/sony/MambaPEFT/blob/main/LICENSE}{Additional-Scan} and the \href{https://github.com/OpenGVLab/VideoMamba/blob/main/LICENSE}{VideoMamba} are licensed under Apache 2.0.

All the datasets included in our study are publicly available, and all the models are publicly available. We would like to state that the contents in the dataset do NOT represent our views or opinions.

\section{Broader Impacts} \label{sec:impacts}

This study presents SSP, which improves the performance of pre-trained Mamba models when fine-tuned on downstream video tasks. Thanks to the reduced parameter count, our research enables deployment of video foundation models on resource-constrained devices, reduced environmental impact of AI training, and rapid adaptation to specialized domains.

\section{Limitations} \label{sec:limitation}

For potential limitations, our method introduces Intra-Frame Gathering (IFG) module and Inter-Frame Spreading (IFS) module to facilitate spatio-temporal contextual information, which brings additional hyperparameters, such as the number of IFSs. However, according to extensive experiments in \cref{sec:IFS_number}, we observe that in most cases, simply setting this parameter to 3 is sufficient. Regarding the internal dimensions of IFG and IFS, as shown in the extensive experiments in \cref{sec:dimension}, we recommend setting the internal dimension of IFG to 384 and the internal dimension of IFS to 256.

% %%%%%%%%%%%%%%%%%%%%%%%%%%%%%%%%%%%%%%%%%%%%%%%%%%%%%%%%%%%%

\end{document}

%% file: nips_checklist.tex
\newpage
\section*{NeurIPS Paper Checklist}

\begin{enumerate}

\item {\bf Claims}
    \item[] Question: Do the main claims made in the abstract and introduction accurately reflect the paper's contributions and scope?
    \item[] Answer: \answerYes{} % Replace by \answerYes{}, \answerNo{}, or \answerNA{}.
    \item[] Justification: We propose SSP, a state space prompting method for video understanding. The main contributions of SSP (i.e., the fine tuning effectiveness and the parameter efficiency) are claimed in both the abstract and introduction accurately.
    \item[] Guidelines:
    \begin{itemize}
        \item The answer NA means that the abstract and introduction do not include the claims made in the paper.
        \item The abstract and/or introduction should clearly state the claims made, including the contributions made in the paper and important assumptions and limitations. A No or NA answer to this question will not be perceived well by the reviewers. 
        \item The claims made should match theoretical and experimental results, and reflect how much the results can be expected to generalize to other settings. 
        \item It is fine to include aspirational goals as motivation as long as it is clear that these goals are not attained by the paper. 
    \end{itemize}

\item {\bf Limitations}
    \item[] Question: Does the paper discuss the limitations of the work performed by the authors?
    \item[] Answer: \answerYes{} % Replace by \answerYes{}, \answerNo{}, or \answerNA{}.
    \item[] Justification: We discuss the limitation in \cref{sec:limitation}.
    \item[] Guidelines:
    \begin{itemize}
        \item The answer NA means that the paper has no limitation while the answer No means that the paper has limitations, but those are not discussed in the paper. 
        \item The authors are encouraged to create a separate "Limitations" section in their paper.
        \item The paper should point out any strong assumptions and how robust the results are to violations of these assumptions (e.g., independence assumptions, noiseless settings, model well-specification, asymptotic approximations only holding locally). The authors should reflect on how these assumptions might be violated in practice and what the implications would be.
        \item The authors should reflect on the scope of the claims made, e.g., if the approach was only tested on a few datasets or with a few runs. In general, empirical results often depend on implicit assumptions, which should be articulated.
        \item The authors should reflect on the factors that influence the performance of the approach. For example, a facial recognition algorithm may perform poorly when image resolution is low or images are taken in low lighting. Or a speech-to-text system might not be used reliably to provide closed captions for online lectures because it fails to handle technical jargon.
        \item The authors should discuss the computational efficiency of the proposed algorithms and how they scale with dataset size.
        \item If applicable, the authors should discuss possible limitations of their approach to address problems of privacy and fairness.
        \item While the authors might fear that complete honesty about limitations might be used by reviewers as grounds for rejection, a worse outcome might be that reviewers discover limitations that aren't acknowledged in the paper. The authors should use their best judgment and recognize that individual actions in favor of transparency play an important role in developing norms that preserve the integrity of the community. Reviewers will be specifically instructed to not penalize honesty concerning limitations.
    \end{itemize}

\item {\bf Theory assumptions and proofs}
    \item[] Question: For each theoretical result, does the paper provide the full set of assumptions and a complete (and correct) proof?
    \item[] Answer: \answerYes{} % Replace by \answerYes{}, \answerNo{}, or \answerNA{}.
    \item[] Justification: All theorems used in the paper are properly referenced.
    \item[] Guidelines:
    \begin{itemize}
        \item The answer NA means that the paper does not include theoretical results. 
        \item All the theorems, formulas, and proofs in the paper should be numbered and cross-referenced.
        \item All assumptions should be clearly stated or referenced in the statement of any theorems.
        \item The proofs can either appear in the main paper or the supplemental material, but if they appear in the supplemental material, the authors are encouraged to provide a short proof sketch to provide intuition. 
        \item Inversely, any informal proof provided in the core of the paper should be complemented by formal proofs provided in appendix or supplemental material.
        \item Theorems and Lemmas that the proof relies upon should be properly referenced. 
    \end{itemize}

    \item {\bf Experimental result reproducibility}
    \item[] Question: Does the paper fully disclose all the information needed to reproduce the main experimental results of the paper to the extent that it affects the main claims and/or conclusions of the paper (regardless of whether the code and data are provided or not)?
    \item[] Answer: \answerYes{} % Replace by \answerYes{}, \answerNo{}, or \answerNA{}.
    \item[] Justification: We claim reproducibility in section \cref{sec:experiments}. Our code will be publicly available after acceptance.
    \item[] Guidelines:
    \begin{itemize}
        \item The answer NA means that the paper does not include experiments.
        \item If the paper includes experiments, a No answer to this question will not be perceived well by the reviewers: Making the paper reproducible is important, regardless of whether the code and data are provided or not.
        \item If the contribution is a dataset and/or model, the authors should describe the steps taken to make their results reproducible or verifiable. 
        \item Depending on the contribution, reproducibility can be accomplished in various ways. For example, if the contribution is a novel architecture, describing the architecture fully might suffice, or if the contribution is a specific model and empirical evaluation, it may be necessary to either make it possible for others to replicate the model with the same dataset, or provide access to the model. In general. releasing code and data is often one good way to accomplish this, but reproducibility can also be provided via detailed instructions for how to replicate the results, access to a hosted model (e.g., in the case of a large language model), releasing of a model checkpoint, or other means that are appropriate to the research performed.
        \item While NeurIPS does not require releasing code, the conference does require all submissions to provide some reasonable avenue for reproducibility, which may depend on the nature of the contribution. For example
        \begin{enumerate}
            \item If the contribution is primarily a new algorithm, the paper should make it clear how to reproduce that algorithm.
            \item If the contribution is primarily a new model architecture, the paper should describe the architecture clearly and fully.
            \item If the contribution is a new model (e.g., a large language model), then there should either be a way to access this model for reproducing the results or a way to reproduce the model (e.g., with an open-source dataset or instructions for how to construct the dataset).
            \item We recognize that reproducibility may be tricky in some cases, in which case authors are welcome to describe the particular way they provide for reproducibility. In the case of closed-source models, it may be that access to the model is limited in some way (e.g., to registered users), but it should be possible for other researchers to have some path to reproducing or verifying the results.
        \end{enumerate}
    \end{itemize}

\item {\bf Open access to data and code}
    \item[] Question: Does the paper provide open access to the data and code, with sufficient instructions to faithfully reproduce the main experimental results, as described in supplemental material?
    \item[] Answer: \answerYes{} % Replace by \answerYes{}, \answerNo{}, or \answerNA{}.
    \item[] Justification: We claim reproducibility in \cref{sec:experiments}. All the datasets included in our study are publicly available. Our code will be publicly available after acceptance. The publicly available code should be adequate to replicate the primary experimental results.
    \item[] Guidelines:
    \begin{itemize}
        \item The answer NA means that paper does not include experiments requiring code.
        \item Please see the NeurIPS code and data submission guidelines (\url{https://nips.cc/public/guides/CodeSubmissionPolicy}) for more details.
        \item While we encourage the release of code and data, we understand that this might not be possible, so “No” is an acceptable answer. Papers cannot be rejected simply for not including code, unless this is central to the contribution (e.g., for a new open-source benchmark).
        \item The instructions should contain the exact command and environment needed to run to reproduce the results. See the NeurIPS code and data submission guidelines (\url{https://nips.cc/public/guides/CodeSubmissionPolicy}) for more details.
        \item The authors should provide instructions on data access and preparation, including how to access the raw data, preprocessed data, intermediate data, and generated data, etc.
        \item The authors should provide scripts to reproduce all experimental results for the new proposed method and baselines. If only a subset of experiments are reproducible, they should state which ones are omitted from the script and why.
        \item At submission time, to preserve anonymity, the authors should release anonymized versions (if applicable).
        \item Providing as much information as possible in supplemental material (appended to the paper) is recommended, but including URLs to data and code is permitted.
    \end{itemize}

\item {\bf Experimental setting/details}
    \item[] Question: Does the paper specify all the training and test details (e.g., data splits, hyperparameters, how they were chosen, type of optimizer, etc.) necessary to understand the results?
    \item[] Answer: \answerYes{} % Replace by \answerYes{}, \answerNo{}, or \answerNA{}.
    \item[] Justification: We specify experimental and implementation details in \cref{sec:experiments}.
    \item[] Guidelines:
    \begin{itemize}
        \item The answer NA means that the paper does not include experiments.
        \item The experimental setting should be presented in the core of the paper to a level of detail that is necessary to appreciate the results and make sense of them.
        \item The full details can be provided either with the code, in appendix, or as supplemental material.
    \end{itemize}

\item {\bf Experiment statistical significance}
    \item[] Question: Does the paper report error bars suitably and correctly defined or other appropriate information about the statistical significance of the experiments?
    \item[] Answer: \answerNo{} % Replace by \answerYes{}, \answerNo{}, or \answerNA{}.
    \item[] Justification: Considering the large size of the datasets, the experiments are too expensive to repeat many
 times. Additionally, by fixing random seeds and releasing our code after publication, our main experimental results are reproducible.
    \item[] Guidelines:
    \begin{itemize}
        \item The answer NA means that the paper does not include experiments.
        \item The authors should answer "Yes" if the results are accompanied by error bars, confidence intervals, or statistical significance tests, at least for the experiments that support the main claims of the paper.
        \item The factors of variability that the error bars are capturing should be clearly stated (for example, train/test split, initialization, random drawing of some parameter, or overall run with given experimental conditions).
        \item The method for calculating the error bars should be explained (closed form formula, call to a library function, bootstrap, etc.)
        \item The assumptions made should be given (e.g., Normally distributed errors).
        \item It should be clear whether the error bar is the standard deviation or the standard error of the mean.
        \item It is OK to report 1-sigma error bars, but one should state it. The authors should preferably report a 2-sigma error bar than state that they have a 96\% CI, if the hypothesis of Normality of errors is not verified.
        \item For asymmetric distributions, the authors should be careful not to show in tables or figures symmetric error bars that would yield results that are out of range (e.g. negative error rates).
        \item If error bars are reported in tables or plots, The authors should explain in the text how they were calculated and reference the corresponding figures or tables in the text.
    \end{itemize}

\item {\bf Experiments compute resources}
    \item[] Question: For each experiment, does the paper provide sufficient information on the computer resources (type of compute workers, memory, time of execution) needed to reproduce the experiments?
    \item[] Answer: \answerYes{} % Replace by \answerYes{}, \answerNo{}, or \answerNA{}.
    \item[] Justification: The compute resources are provided in \cref{sec:implementation}.
    \item[] Guidelines:
    \begin{itemize}
        \item The answer NA means that the paper does not include experiments.
        \item The paper should indicate the type of compute workers CPU or GPU, internal cluster, or cloud provider, including relevant memory and storage.
        \item The paper should provide the amount of compute required for each of the individual experimental runs as well as estimate the total compute. 
        \item The paper should disclose whether the full research project required more compute than the experiments reported in the paper (e.g., preliminary or failed experiments that didn't make it into the paper). 
    \end{itemize}
    
\item {\bf Code of ethics}
    \item[] Question: Does the research conducted in the paper conform, in every respect, with the NeurIPS Code of Ethics \url{https://neurips.cc/public/EthicsGuidelines}?
    \item[] Answer: \answerYes{} % Replace by \answerYes{}, \answerNo{}, or \answerNA{}.
    \item[] Justification: This paper complies with the NeurIPS Code of Ethics.
    \item[] Guidelines:
    \begin{itemize}
        \item The answer NA means that the authors have not reviewed the NeurIPS Code of Ethics.
        \item If the authors answer No, they should explain the special circumstances that require a deviation from the Code of Ethics.
        \item The authors should make sure to preserve anonymity (e.g., if there is a special consideration due to laws or regulations in their jurisdiction).
    \end{itemize}

\item {\bf Broader impacts}
    \item[] Question: Does the paper discuss both potential positive societal impacts and negative societal impacts of the work performed?
    \item[] Answer: \answerYes{}. % Replace by \answerYes{}, \answerNo{}, or \answerNA{}.
    \item[] Justification: The \textbf{social impact} of our research is discussed in \cref{sec:impacts}.
    \item[] Guidelines:
    \begin{itemize}
        \item The answer NA means that there is no societal impact of the work performed.
        \item If the authors answer NA or No, they should explain why their work has no societal impact or why the paper does not address societal impact.
        \item Examples of negative societal impacts include potential malicious or unintended uses (e.g., disinformation, generating fake profiles, surveillance), fairness considerations (e.g., deployment of technologies that could make decisions that unfairly impact specific groups), privacy considerations, and security considerations.
        \item The conference expects that many papers will be foundational research and not tied to particular applications, let alone deployments. However, if there is a direct path to any negative applications, the authors should point it out. For example, it is legitimate to point out that an improvement in the quality of generative models could be used to generate deepfakes for disinformation. On the other hand, it is not needed to point out that a generic algorithm for optimizing neural networks could enable people to train models that generate Deepfakes faster.
        \item The authors should consider possible harms that could arise when the technology is being used as intended and functioning correctly, harms that could arise when the technology is being used as intended but gives incorrect results, and harms following from (intentional or unintentional) misuse of the technology.
        \item If there are negative societal impacts, the authors could also discuss possible mitigation strategies (e.g., gated release of models, providing defenses in addition to attacks, mechanisms for monitoring misuse, mechanisms to monitor how a system learns from feedback over time, improving the efficiency and accessibility of ML).
    \end{itemize}
    
\item {\bf Safeguards}
    \item[] Question: Does the paper describe safeguards that have been put in place for responsible release of data or models that have a high risk for misuse (e.g., pretrained language models, image generators, or scraped datasets)?
    \item[] Answer: \answerNA{} % Replace by \answerYes{}, \answerNo{}, or \answerNA{}.
    \item[] Justification: The paper poses no such risks.
    \item[] Guidelines:
    \begin{itemize}
        \item The answer NA means that the paper poses no such risks.
        \item Released models that have a high risk for misuse or dual-use should be released with necessary safeguards to allow for controlled use of the model, for example by requiring that users adhere to usage guidelines or restrictions to access the model or implementing safety filters. 
        \item Datasets that have been scraped from the Internet could pose safety risks. The authors should describe how they avoided releasing unsafe images.
        \item We recognize that providing effective safeguards is challenging, and many papers do not require this, but we encourage authors to take this into account and make a best faith effort.
    \end{itemize}

\item {\bf Licenses for existing assets}
    \item[] Question: Are the creators or original owners of assets (e.g., code, data, models), used in the paper, properly credited and are the license and terms of use explicitly mentioned and properly respected?
    \item[] Answer: \answerYes{} % Replace by \answerYes{}, \answerNo{}, or \answerNA{}.
    \item[] Justification: We show related asset license and consent to our work in \cref{sec:asset}.
    \item[] Guidelines:
    \begin{itemize}
        \item The answer NA means that the paper does not use existing assets.
        \item The authors should cite the original paper that produced the code package or dataset.
        \item The authors should state which version of the asset is used and, if possible, include a URL.
        \item The name of the license (e.g., CC-BY 4.0) should be included for each asset.
        \item For scraped data from a particular source (e.g., website), the copyright and terms of service of that source should be provided.
        \item If assets are released, the license, copyright information, and terms of use in the package should be provided. For popular datasets, \url{paperswithcode.com/datasets} has curated licenses for some datasets. Their licensing guide can help determine the license of a dataset.
        \item For existing datasets that are re-packaged, both the original license and the license of the derived asset (if it has changed) should be provided.
        \item If this information is not available online, the authors are encouraged to reach out to the asset's creators.
    \end{itemize}

\item {\bf New assets}
    \item[] Question: Are new assets introduced in the paper well documented and is the documentation provided alongside the assets?
    \item[] Answer: \answerNA{} % Replace by \answerYes{}, \answerNo{}, or \answerNA{}.
    \item[] Justification: The paper does not release new assets.
    \item[] Guidelines:
    \begin{itemize}
        \item The answer NA means that the paper does not release new assets.
        \item Researchers should communicate the details of the dataset/code/model as part of their submissions via structured templates. This includes details about training, license, limitations, etc. 
        \item The paper should discuss whether and how consent was obtained from people whose asset is used.
        \item At submission time, remember to anonymize your assets (if applicable). You can either create an anonymized URL or include an anonymized zip file.
    \end{itemize}

\item {\bf Crowdsourcing and research with human subjects}
    \item[] Question: For crowdsourcing experiments and research with human subjects, does the paper include the full text of instructions given to participants and screenshots, if applicable, as well as details about compensation (if any)? 
    \item[] Answer: \answerNA{} % Replace by \answerYes{}, \answerNo{}, or \answerNA{}.
    \item[] Justification: The paper does not involve crowdsourcing nor research with human subjects.
    \item[] Guidelines:
    \begin{itemize}
        \item The answer NA means that the paper does not involve crowdsourcing nor research with human subjects.
        \item Including this information in the supplemental material is fine, but if the main contribution of the paper involves human subjects, then as much detail as possible should be included in the main paper. 
        \item According to the NeurIPS Code of Ethics, workers involved in data collection, curation, or other labor should be paid at least the minimum wage in the country of the data collector. 
    \end{itemize}

\item {\bf Institutional review board (IRB) approvals or equivalent for research with human subjects}
    \item[] Question: Does the paper describe potential risks incurred by study participants, whether such risks were disclosed to the subjects, and whether Institutional Review Board (IRB) approvals (or an equivalent approval/review based on the requirements of your country or institution) were obtained?
    \item[] Answer: \answerNA{} % Replace by \answerYes{}, \answerNo{}, or \answerNA{}.
    \item[] Justification: The paper does not involve crowdsourcing nor research with human subjects.
    \item[] Guidelines:
    \begin{itemize}
        \item The answer NA means that the paper does not involve crowdsourcing nor research with human subjects.
        \item Depending on the country in which research is conducted, IRB approval (or equivalent) may be required for any human subjects research. If you obtained IRB approval, you should clearly state this in the paper. 
        \item We recognize that the procedures for this may vary significantly between institutions and locations, and we expect authors to adhere to the NeurIPS Code of Ethics and the guidelines for their institution. 
        \item For initial submissions, do not include any information that would break anonymity (if applicable), such as the institution conducting the review.
    \end{itemize}

\item {\bf Declaration of LLM usage}
    \item[] Question: Does the paper describe the usage of LLMs if it is an important, original, or non-standard component of the core methods in this research? Note that if the LLM is used only for writing, editing, or formatting purposes and does not impact the core methodology, scientific rigorousness, or originality of the research, declaration is not required.
    %this research? 
    \item[] Answer: \answerNA{} % Replace by \answerYes{}, \answerNo{}, or \answerNA{}.
    \item[] Justification: The core method development in this research does not involve LLMs as any important, original, or non-standard components.
    \item[] Guidelines:
    \begin{itemize}
        \item The answer NA means that the core method development in this research does not involve LLMs as any important, original, or non-standard components.
        \item Please refer to our LLM policy (\url{https://neurips.cc/Conferences/2025/LLM}) for what should or should not be described.
    \end{itemize}

\end{enumerate}